\pdfoutput=1

\documentclass[10pt,twocolumn,letterpaper]{article}

\usepackage{cvpr}
\usepackage{times}
\usepackage{epsfig}
\usepackage{graphicx}
\usepackage{amsmath}
\usepackage{amssymb}

\usepackage{times}
\usepackage{graphicx}
\usepackage{amsmath}
\usepackage{amssymb}

\usepackage{mathrsfs}
\usepackage{amsfonts}
\usepackage{amsthm}

\usepackage[lined,boxed]{algorithm2e}

\usepackage{bm}
\usepackage{setspace}

\newcommand{\ninesevenhao}{\fontsize{6pt}{6pt}\selectfont}

\usepackage[tight,rm]{subfigure}
\usepackage{xspace,algorithmic,algorithm,bibspacing}
\graphicspath{{./}{./figs/}{./caltech/}}

\newcommand{\comment}[1]{}

\renewcommand{\comment}[1]{ {\color{red}{COMMENTS}:}{\color{red}{ #1} }}

\def\bp{ {\bf p } }

\def\bM{ {\bf M } }

\usepackage{caption}
\usepackage[compress,nosort]{cite}

\cvprfinalcopy %

\begin{document}

\title{Non-sparse Linear Representations for Visual Tracking \\
with Online Reservoir Metric Learning}

\author{Xi Li, Chunhua Shen, Qinfeng  Shi, Anthony Dick,
Anton van den Hengel\\
Australian Centre for Visual Technologies, The University of Adelaide,
SA 5005, Australia\\
}

\maketitle

\begin{abstract}

    Most sparse linear representation-based trackers need to solve a
    computationally expensive $\ell_{1}$-regularized optimization
    problem.
    To address this problem, we propose a visual tracker based on
    {\em non-sparse} linear representations, which admit an efficient
    closed-form solution without sacrificing accuracy.
    Moreover, in order to capture the correlation information between
    different feature dimensions, we learn a Mahalanobis distance
    metric in an online fashion and incorporate the learned metric
    into the optimization problem for obtaining the linear representation.
    We show that online metric learning using proximity comparison
    significantly improves the robustness of the
    tracking, especially on those sequences exhibiting drastic
    appearance changes.
    Furthermore,
    in order  to prevent the
    unbounded growth in the number of training samples for the metric
    learning,
    we design a time-weighted reservoir sampling method
    to %
    maintain and update limited-sized
    foreground and background sample buffers for balancing
    sample diversity and adaptability.
    Experimental results
    on challenging videos demonstrate the effectiveness and robustness
    of the proposed tracker.

\end{abstract}

\section{Introduction}

Robust visual tracking is an important problem in computer vision.
In recent years, steady improvements have been made to the speed,
accuracy and robustness of tracking techniques. A crucial factor in many
of these improvements has been the construction and optimization of
object appearance models ({\em e.g.},
\cite{Limy-Ross17,Meo-Ling-ICCV09,
Kwon-Lee-CVPR2010,Li-Shen-Shi-cvpr2011,harestruck_iccv2011,li2007robust,li2011graph,li2010robust,
Generalized2010Shen}).
Among these models, the linear representation, in which the object is
represented as a linear combination of basis samples, has proved to be
a simple yet effective choice.
For example, Mei and Ling~\cite{Meo-Ling-ICCV09} propose a tracker
based on a sparse linear representation
which  solves an $\ell_{1}$-regularized optimization problem.
With the sparsity constraint, this tracker obtains a sparse regression solution
that can adaptively select a small number of
relevant templates to optimally approximate the given test samples.
The drawback is its  expensive computation due to the need of solving
an $\ell_{1}$-norm convex problem.
    To speed up the tracking,
    Li \emph{et al.}~\cite{Li-Shen-Shi-cvpr2011}
    propose to  approximately solve the sparsity
    optimization problem using orthogonal matching pursuit (OMP).
Recently,
    research has revealed that the $ \ell_1 $-norm induced sparsity
    does not in general help improve the accuracy of image classification;
    and non-sparse representation based methods are typically orders of
    magnitudes faster than the sparse representation based ones with
    competitive and sometimes even better accuracy
    \cite{Shi-Eriksson-Hengel-Shen-cvpr2011,rigamonti2011sparse,FaceICCV2011}.

    Inspired by these findings, here we propose a {\em non-sparse}
    linear representation based visual tracker. The proposed tracker
    can be implemented by solving a least-square problem, which admits
    an extremely simple and efficient closed-form solution.  To date,
    linear representation based trackers
    \cite{Meo-Ling-ICCV09,Li-Shen-Shi-cvpr2011} have built
    linear regressors that are defined on independent feature dimensions
    (mutually independent raw pixels in both \cite{Meo-Ling-ICCV09}
    and \cite{Li-Shen-Shi-cvpr2011}). In other words, the correlation
    information between different feature dimensions is not exploited. We argue that
    this correlation information is important in tracking.
    To address this problem,
    we learn a Mahalanobis distance metric and incorporate it into the
    optimization of the linear representation.

    Metric learning
    has emerged as a useful tool for many applications.
    For example, in \cite{weinberger2006distance,Shen2009PSD}, a
    Mahalanobis distance metric is learned using positive semidefinite
    programming. 
    Discriminative metric learning has also been successfully applied
    to visual tracking
    \cite{wang2010discriminative,jiang2011adaptive}.
    These works learn a distance
    metric  mainly for object matching across adjacent frames, and
    the tracking is not carried out in the framework of linear
    representations.
    In this work, we learn a distance metric using proximity
    comparison for linear representation based tracking.
    The learning strategy is adapted from  the online metric learning for
    image retrieval of  Chechik \emph{et al.} \cite{chechik2010large}.
    There,  it has been shown that the online
    learning procedure is efficient and capable for large-scale
    learning. Nevertheless, it is not designed for dealing with time-varying
    data stream such as in real-time visual tracking.

        Visual tracking is a time-varying process which deals with a
        dynamic stream data in an online manner. Due to memory limit,
        it is often impractical for trackers to store all the stream data.
        Furthermore, visual tracking in the current frame usually
        relies more on recently received samples than old samples due
        to its temporal coherence property.  Therefore, it is
        necessary for trackers to maintain and update limited-sized
        data buffers for balancing between sample diversity and
        adaptability.  To address this issue, we propose to use
        reservoir sampling \cite{vitter1985random, zhaoICML2011} for
        sequential random sampling. The conventional reservoir
        sampling in \cite{vitter1985random, zhaoICML2011} can only
        accomplish the task of uniform random sampling, which ignores
        the importance variance among samples.  We therefore need a
        time-weighted reservoir sampling.

        In summary, we propose a robust tracker that is based on
        metric-weighted linear representations and time-weighted
        reservoir sampling. Our main contributions are as follows. 
\begin{enumerate}
\setlength{\itemsep}{-1.2mm}
\item

    We propose an online discriminative linear representation for
    visual tracking. The metric-weighted least-square optimization
    problem admits a closed-form solution, which significantly
    improves tracking efficiency.  We also demonstrate that, with the
    emergence of new data, the closed-form solution can be efficiently
    updated by a sequence of simple matrix operations.

\item
 To further improve the discriminative capability of the
linear representation for distinguishing foreground and background,
we present an online Mahalanobis distance metric learning method and
incorporate the learned metric into the optimization problem for obtaining
a discriminative linear representation.
The learned metric can effectively capture the correlation
information between different feature dimensions. Such correlation
information plays an important role in robust object/non-object classification.

\item
    
    To allow for real-time applications,
we design a time-weighted reservoir sampling method to maintain and update
limited-sized sample buffers for balancing
between sample diversity and adaptability in the metric learning
procedure.
With the theory of
\cite{Kolonko04sequentialreservoir,efraimidis2006weighted},
larger weights are assigned to those recently received samples,
which is particularly important for tracking.
To our knowledge, {\em it is the first time that reservoir sampling is used
in an  online metric learning setting that is tailored for robust
visual tracking.}

\end{enumerate}

\section{The proposed visual tracker}

    In this section, we describe the novel aspects of the proposed visual tracker:
\begin{enumerate} \itemsep = -2pt
\item
 Object state estimation. This is implemented by an online
metric-weighted optimization, as described in Section~\ref{sec:malso};
\item
 Metric update using the online metric-weighted optimization in
    response to changing foreground and background, as described in
    Section~\ref{sec:oml};
\item
 Sample update used for object representation based on reservoir
 sampling, as
    described in Section~\ref{sec:twrs}.
\end{enumerate}

\subsection{Online metric-weighted linear representation}
\label{sec:malso}

    To effectively characterize dynamic appearance variations during
    tracking, an object is associated with an appearance subspace
    spanned by a set of basis samples, which encode the distribution
    of the object appearance. Therefore, the problem of visual
    tracking is converted to that of linear representation and
    reconstruction. As a result, the sample-to-subspace distance
    (e.g., linear reconstruction error)  can be used for evaluating
    the likelihood of a test sample belonging to the object
    appearance. However, the conventional linear representations
    (e.g., used in~\cite{Meo-Ling-ICCV09, Li-Shen-Shi-cvpr2011})
    ignore the correlation information between feature dimensions. Due
    to the influence of complicated appearance variations, the
    correlation across feature dimensions usually differs greatly
    during tracking. In order to address this problem, we propose a
    metric-weighted linear representation based on solving a
    metric-weighted optimization problem under a learned distance
    metric. Consequently, the proposed linear representation is
    capable of capturing the varying correlation information between
    feature dimensions.

More specifically, given a set of basis samples $\mathbf{P} =
(\mathbf{p_{i}})_{i=1}^{N} \in \mathcal{R}^{d\times N}$ and a test
sample $\mathbf{y} \in \mathcal{R}^{d\times 1}$, we aim to discover
a linear combination of $\mathbf{P}$ to optimally approximate the test
sample $\mathbf{y}$ by solving the following optimization problem: \vspace{-0.2cm}
\begin{equation}
 \underset{\mathbf{x}}{\min} \thinspace g(\mathbf{x}; \mathbf{M},
\mathbf{P}, \mathbf{y})= \underset{\mathbf{x}}{\min} \thinspace
(\mathbf{y}-\mathbf{P}\mathbf{x})^{T}\mathbf{M}
(\mathbf{y}-\mathbf{P}\mathbf{x}),
\label{eq:linear_regression_metric} \vspace{-0.2cm}
\end{equation}
where $\mathbf{x} \in \mathcal{R}^{N \times 1}$ and
$\mathbf{M}$ is a symmetric distance metric matrix.
The optimization problem~\eqref{eq:linear_regression_metric} is a
weighted linear regression problem whose analytical solution can
be directly computed as: \vspace{-0.2cm}
\begin{equation}
\mathbf{x}^{\ast} =
(\mathbf{P}^{T}\mathbf{M}\mathbf{P})^{-1}\mathbf{P}^{T}\mathbf{M}\mathbf{y}.
\label{eq:regression_optimal_solution} \vspace{-0.2cm}
\end{equation}
If $\mathbf{P}^{T}\mathbf{M}\mathbf{P}$ is a singular matrix, we
directly use its pseudoinverse to compute $\mathbf{x}^{\ast}$.
The main computational time of
Equ.~\eqref{eq:regression_optimal_solution} is spent on the calculation
of
$(\mathbf{P}^{T}\mathbf{M}\mathbf{P})^{-1}$.
For computational efficiency, we need to incrementally or decrementally
update the inverse when $\mathbf{P}$ is expanded or reduced with one
column under the same
metric $\mathbf{M}$.
Let $\mathbf{P_{n}} = (\mathbf{P} \thickspace \Delta \mathbf{p})$
denote the expanded matrix of $\mathbf{P}$.
Clearly, the following relation holds: 
\[
(\mathbf{P_{n}})^{T}\mathbf{M}\mathbf{P_{n}}
=\left(
\begin{matrix}
\mathbf{P}^{T}\mathbf{M}\mathbf{P} & \mathbf{P}^{T}\mathbf{M} \Delta
\mathbf{p}\\
(\Delta \mathbf{p})^{T}\mathbf{M} \mathbf{P} & (\Delta
\mathbf{p})^{T}\mathbf{M} \Delta \mathbf{p}
\end{matrix}
\right). 
\]
For simplicity, let $\mathbf{H}=(\mathbf{P}^{T}\mathbf{M}\mathbf{P})^{-1}$,
$\mathbf{c} = \mathbf{P}^{T}\mathbf{M} \Delta \mathbf{p}$, and $r =
(\Delta \mathbf{p})^{T}\mathbf{M} \Delta \mathbf{p}$.
Since $\mathbf{M}$ is a symmetric matrix,
$\mathbf{c}^{T} = (\Delta \mathbf{p})^{T}\mathbf{M} \mathbf{P}$.
According to the theory of matrix
computation~\cite{jennings1992matrix}, the corresponding inverse of
 $(\mathbf{P_{n}})^{T}\mathbf{M}\mathbf{P_{n}}$ can be computed
as: 
\begin{equation}
((\mathbf{P_{n}})^{T}\mathbf{M}\mathbf{P_{n}})^{-1}
=\left(
\begin{matrix}
\mathbf{H}+\frac{\mathbf{H}\mathbf{c}\mathbf{c}^{T}\mathbf{H}}{r-\mathbf{c}^{T}\mathbf{H}\mathbf{c}}
& -\frac{\mathbf{H}\mathbf{c}}{r-\mathbf{c}^{T}\mathbf{H}\mathbf{c}}\\
-\frac{\mathbf{c}^{T}\mathbf{H}}{r-\mathbf{c}^{T}\mathbf{H}\mathbf{c}}
& \frac{1}{r-\mathbf{c}^{T}\mathbf{H}\mathbf{c}}
\end{matrix}
\right).
\label{eq:incremental} 
\end{equation}
Similarly, let
$\mathbf{P_{o}}$ denote the reduced matrix of $\mathbf{P}$ after
removing the $i$-th column such that $1\leq i \leq N$.
Based on~\cite{jennings1992matrix}, the corresponding inverse of
$(\mathbf{P_{o}})^{T}\mathbf{M}\mathbf{P_{o}}$ can be computed
as: \vspace{-0.082cm}
\begin{equation}
((\mathbf{P_{o}})^{T}\mathbf{M}\mathbf{P_{o}})^{-1}
=\mathbf{H}(\mathcal{I}_{i},\mathcal{I}_{i})- \frac{\mathbf{H}(\mathcal{I}_{i},
i)\mathbf{H}(i,\mathcal{I}_{i})}{\mathbf{H}(i,i)},\\
\label{eq:decremental} \vspace{-0.082cm}
\end{equation}
where $\mathcal{I}_{i}=\{1, 2, \ldots, N\}\backslash{\{i\}}$ stands for
the index set except $i$.
For adapting to object appearance changes, it is necessary for
trackers to replace an old sample from the sample buffer with a new
sample.
In essence, the replacement operation can be decomposed into two
stages: 1) old sample removal; and 2) new sample arrival.
As a matter of fact, 1) and 2) correspond to the decremental and
incremental cases, respectively.
Given $\mathbf{H}=(\mathbf{P}^{T}\mathbf{M}\mathbf{P})^{-1}$, we first
compute the decremental inverse
$((\mathbf{P_{o}})^{T}\mathbf{M}\mathbf{P_{o}})^{-1}$
according to Equ.~\eqref{eq:decremental},
and then calculate the
incremental inverse $((\mathbf{P_{o}} \thickspace \Delta
\mathbf{p})^{T}\mathbf{M}(\mathbf{P_{o}} \thickspace \Delta
\mathbf{p}))^{-1}$ using Equ.~\eqref{eq:incremental}.
For notational simplicity, we let $\mathbf{P'} = (\mathbf{P_{o}} \thickspace \Delta \mathbf{p})$,
$\mathbf{H_{o}}=((\mathbf{P_{o}})^{T}\mathbf{M}\mathbf{P_{o}})^{-1}$,
$\mathbf{c_{o}} = (\mathbf{P_{o}})^{T}\mathbf{M} \Delta \mathbf{p}$,
and $r = (\Delta \mathbf{p})^{T}\mathbf{M} \Delta \mathbf{p}$.
Based on Equ.~\eqref{eq:incremental}, $((\mathbf{P'})^{T}\mathbf{M} \mathbf{P'})^{-1}$
can be computed as:\vspace{-0.02cm}
\begin{equation}
\hspace{-0.1cm}
((\mathbf{P'})^{T}\mathbf{M} \mathbf{P'})^{-1}
\hspace{-0.1cm}= \hspace{-0.1cm}\left(
\begin{matrix}
\mathbf{H_{o}}+\frac{\mathbf{H_{o}}\mathbf{c_{o}}\mathbf{c_{o}}^{T}\mathbf{H_{o}}}{r-\mathbf{c_{o}}^{T}\mathbf{H_{o}}\mathbf{c_{o}}}
& -\frac{\mathbf{H_{o}}\mathbf{c_{o}}}{r-\mathbf{c_{o}}^{T}\mathbf{H_{o}}\mathbf{c_{o}}}\\
-\frac{\mathbf{c_{o}}^{T}\mathbf{H_{o}}}{r-\mathbf{c_{o}}^{T}\mathbf{H_{o}}\mathbf{c_{o}}}
& \frac{1}{r-\mathbf{c_{o}}^{T}\mathbf{H_{o}}\mathbf{c_{o}}}
\end{matrix}
\right)\hspace{-0.5cm}
\label{eq:final_replacement} \vspace{-0.082cm}
\end{equation}
Furthermore, when updated according to
Algorithm~\ref{alg:online_metric_learning}, $\mathbf{M}$ is modified
as a rank-one
addition
such that $\mathbf{M}\longleftarrow \mathbf{M} + \eta
(\mathbf{a}_{-}\mathbf{a}_{-}^{T} -
\mathbf{a}_{+}\mathbf{a}_{+}^{T})$ where $\mathbf{a}_{+} = \mathbf{p} - \mathbf{p}^{+}$ and
$\mathbf{a}_{-} = \mathbf{p} - \mathbf{p}^{-}$ are
two vectors (defined in Equ.~\eqref{eq:triplet_update}) for triplet construction, and
$\eta$ is a step-size factor (defined in Equ.~\eqref{eq:eta_final}).
As a result, the original $\mathbf{P}^{T}\mathbf{M}\mathbf{P}$
becomes $\mathbf{P}^{T}\mathbf{M}\mathbf{P}
+ (\eta\mathbf{P}^{T}\mathbf{a}_{-})(\mathbf{P}^{T}\mathbf{a}_{-})^{T}
+  (-\eta\mathbf{P}^{T}\mathbf{a}_{+})(\mathbf{P}^{T}\mathbf{a}_{+})^{T}$.
When $\mathbf{M}$ is modified by a rank-one addition, the inverse of
$\mathbf{P}^{T}\mathbf{M}\mathbf{P}$ can be easily updated
according to the theory of~\cite{householder1964theory,powell1969theorem}: 
\begin{equation}
(\mathbf{J}+\mathbf{u}\mathbf{v}^{T})^{-1} = \mathbf{J}^{-1} -
\frac{\mathbf{J}^{-1}\mathbf{u}\mathbf{v}^{T}\mathbf{J}^{-1}}{1+\mathbf{v}^{T}\mathbf{J}^{-1}\mathbf{u}}.
\label{eq:rank_one_update} \vspace{-0.02cm}
\end{equation}
Here, $\mathbf{J}=\mathbf{P}^{T}\mathbf{M}\mathbf{P}$, $\mathbf{u}=\eta\mathbf{P}^{T}\mathbf{a}_{-}$ (or $\mathbf{u}=-\eta\mathbf{P}^{T}\mathbf{a}_{+}$),
and $\mathbf{v}=\mathbf{P}^{T}\mathbf{a}_{-}$ (or $\mathbf{v}=\mathbf{P}^{T}\mathbf{a}_{+}$).
The complete procedure of online linear optimization under the metric
$\mathbf{M}$ is summarized in
Algorithm~\ref{alg:online_linear_regression}.

\begin{algorithm}[t]
\begin{minipage}[ctb]{8.5cm}
\footnotesize
\caption{Metric-weighted linear representation}
\label{alg:online_linear_regression}
\KwIn
{
 The current distance metric matrix $\mathbf{M}$, the basis samples
$\mathbf{P} = (\mathbf{p_{i}})_{i=1}^{N} \in \mathcal{R}^{d\times N}$,
any test sample $\mathbf{y} \in \mathcal{R}^{d\times 1}$.
}
\KwOut
{
 The optimal linear representation solution $\mathbf{x}^{\ast}$.
}
\begin{enumerate}\itemsep=-3.8pt
  \item Build the optimization problem in
Equ.~\eqref{eq:linear_regression_metric}:\vspace{-0.15cm}
       \[\underset{\mathbf{x}}{\min} \thinspace g(\mathbf{x};
\mathbf{P}, \mathbf{y})= \underset{\mathbf{x}}{\min} \thinspace
(\mathbf{y}-\mathbf{P}\mathbf{x})^{T}\mathbf{M}
(\mathbf{y}-\mathbf{P}\mathbf{x})
        \vspace{-0.15cm}\]
  \item Compute the optimal solution $\mathbf{x}^{\ast} =
(\mathbf{P}^{T}\mathbf{M}\mathbf{P})^{-1}\mathbf{P}^{T}\mathbf{M}\mathbf{y}$.
When $\mathbf{P}$ is expanded, reduced, or replaced by one column,
the corresponding computation of $(\mathbf{P}^{T}\mathbf{M}\mathbf{P})^{-1}$
        can be efficiently accomplished in an online manner: \vspace{-0.25cm}
        \begin{itemize} \itemsep=-2.8pt
        \item Use Equ.~\eqref{eq:incremental} to compute the
incremental inverse.
        \item Employ Equ.~\eqref{eq:decremental} to calculate the
decremental inverse.
        \item Utilize Equ.~\eqref{eq:final_replacement} to obtain the
replacement inverse.
        \end{itemize}
  \item Update the inverse of $\mathbf{P}^{T}\mathbf{M}\mathbf{P}$ by
Equ.~\eqref{eq:rank_one_update} when $\mathbf{M}$ is modified as
        a rank-one addition in
Algorithm~\ref{alg:online_metric_learning}, and then repeat Steps 1
and 2.
  \item Return the optimal  solution $\mathbf{x}^{\ast}$.
\end{enumerate}
\end{minipage}
\end{algorithm}

Furthermore, visual tracking is typically posed as a binary
classification problem. To address this problem, we need to
simultaneously optimize the following two
objective functions:
$\mathbf{x}^{\ast}_{f} = \arg \min_{\mathbf{x}_{f}}  g(\mathbf{x}_{f};
\mathbf{M}, \mathbf{P}_{f}, \mathbf{y})$ and
$\mathbf{x}^{\ast}_{b} = \arg \min_{\mathbf{x}_{b}}  g(\mathbf{x}_{b};
\mathbf{M}, \mathbf{P}_{b}, \mathbf{y})$,
where $\mathbf{P}_{f}$ and $\mathbf{P}_{b}$ are foreground and
background basis samples, respectively.
Thus, we can define a discriminative criterion for measuring the
similarity of the test sample $\mathbf{y}$ belonging to foreground
class: \vspace{-0.13cm}
\begin{equation}
\mathcal{S}(\mathbf{y}) \hspace{-0.06cm} =
\hspace{-0.06cm}\sigma\left[\exp(-\theta_{f}/\gamma_{f}) -
\rho \exp(-\theta_{b}/\gamma_{b})\right],
\label{eq:particle_liki_model} \vspace{-0.083cm}
\end{equation}
where $\gamma_{f}$ and $\gamma_{b}$ are two scaling factors,
$\theta_{f} $ = $ g(\mathbf{x}_{f}^{\ast}; $ $ \mathbf{M},
\mathbf{P}_{f}, \mathbf{y})$, $\theta_{b} =
g(\mathbf{x}_{b}^{\ast}; \mathbf{M}, \mathbf{P}_{b},
\mathbf{y})$,
$\rho$ is a trade-off control factor, and $\sigma[\cdot]$ is the
sigmoid function.

\begin{algorithm}[t]
\begin{minipage}[ctb]{8.5cm}
\footnotesize
\caption{Online distance metric learning using triplets}
\label{alg:online_metric_learning}
\KwIn
{
 The current distance metric matrix $\mathbf{M}^{k}$ and a new
triplet $(\mathbf{p}, \mathbf{p}^{+}, \mathbf{p}^{-})$.
}

\KwOut
{
 The updated distance metric matrix $\mathbf{M}^{k+1}$.
}
\begin{enumerate}\itemsep=-2.8pt
  \item Calculate $\mathbf{a}_{+} = \mathbf{p} - \mathbf{p}^{+}$ and
$\mathbf{a}_{-} = \mathbf{p} - \mathbf{p}^{-}$
  \item Compute the optimal step length $\eta$ that is formulated as:
  $\eta \hspace{-0.08cm}= \hspace{-0.08cm}\min\left\{C, \max \left\{0,\frac{1 + \mathbf{a}_{+}^{T}\mathbf{M}^{k}\mathbf{a}_{+} -
\mathbf{a}_{-}^{T}\mathbf{M}^{k}\mathbf{a}_{-}}{2\mathbf{a}_{-}^{T}\mathbf{U}\mathbf{a}_{-} \hspace{-0.02cm}- \hspace{-0.02cm}
2\mathbf{a}_{+}^{T}\mathbf{U}\mathbf{a}_{+} \hspace{-0.02cm}- \hspace{-0.02cm}\|\mathbf{U}\|_{F}^{2}}\right\}\hspace{-0.1cm}\right\}$
  with $\mathbf{U}$ being $\mathbf{a}_{-}\mathbf{a}_{-}^{T} -
\mathbf{a}_{+}\mathbf{a}_{+}^{T}$.
  \item $\mathbf{M}^{k+1} \leftarrow \mathbf{M}^{k} + \eta
(\mathbf{a}_{-}\mathbf{a}_{-}^{T} -
\mathbf{a}_{+}\mathbf{a}_{+}^{T})$.
\end{enumerate}
\end{minipage}
\end{algorithm}

\subsection{Online metric learning using proximity comparison}
\label{sec:oml}

To efficiently compute the linear representation solution
in Equ.~\eqref{eq:regression_optimal_solution}, we need to update the
quadratic Mahalanobis
 distance metric in an online manner. Motivated by this,
we propose an online metric learning scheme by solving a max-margin
optimization problem using triplets.

   Suppose that we have a set of triplets
   $\{(\mathbf{p}, \mathbf{p}^{+},\mathbf{p}^{-})\}$
    with
    $ \mathbf{p},
    \mathbf{p}^{+},
    \mathbf{p}^{-} \in \mathcal{R}^{d}
    $.  These triplets encode the proximity comparison information.
   Without loss of generality, let us assume that the distance between $  \bf p $
   and $ {\bf p}^+  $ is smaller than the distance between
   $ \bf p $ and $ {\bf p}^-  $.

The Mahalanobis distance under  metric $ \bf M$ is defined as:
\vspace{-0.153cm}
\begin{equation}
D_{\mathbf{M}}(\mathbf{p}, \mathbf{q}) = (\mathbf{p}
-\mathbf{q})^{T}\mathbf{M}(\mathbf{p}-\mathbf{q}). \vspace{-0.126cm}
\end{equation}
Clearly, $ \bf M $ must be a symmetric and positive semidefinite
matrix. It is equivalent to learn a projection matrix $ \bf L $ such
that $ {\bf M} = {\bf L} {\bf L}^T$.
In practice,
   we generate the triplets set as:
   $ \bp $ and $ \bp^+$ belong to the same class
   and $ \bp $ and $ \bp^-$ belong to different classes.
So we want the constraints
$ D_\bM ( \bp, \bp^+ ) < D_\bM ( \bp, \bp^- ) $ to be satisfied as
well as possible.
By putting it into a large-margin learning framework, and using the
soft-margin hinge loss,
the loss function for each triplet is: \vspace{-0.016cm}
\begin{equation}
l_{\mathbf{M}}(\mathbf{p}, \mathbf{p}^{+}, \mathbf{p}^{-}) =
       \max\{0, 1 + D_{\mathbf{M}}(\mathbf{p}, \mathbf{p}^{+}) -
D_{\mathbf{M}}(\mathbf{p}, \mathbf{p}^{-})\}. \vspace{-0.01cm}
\label{eq:local_hinge_loss}
\end{equation}
To obtain the optimal distance metric matrix $\mathbf{M}$, we need to
minimize the global loss $L_{\mathbf{M}}$
that takes the sum of hinge losses~\eqref{eq:local_hinge_loss} over
all possible triplets from the training
set: \vspace{-0.21cm}
\begin{equation}
L_{\mathbf{M}} = \underset{(\mathbf{p}, \mathbf{p}^{+},
\mathbf{p}^{-})\in \mathcal{Q}}{\sum} l_{\mathbf{M}}(\mathbf{p},
\mathbf{p}^{+}, \mathbf{p}^{-}), \vspace{-0.21cm}
\label{eq:total_loss}
\end{equation}
where $\mathcal{Q}$ is the triplet set.
   To sequentially optimize the above objective function $L_{\mathbf{M}}$
   in an online fashion,
   we design an iterative algorithm to solve the following
   convex problem: \vspace{-0.22cm}
\begin{equation}
\begin{array}{l}
\mathbf{M}^{k+1} = \underset{\mathbf{M}}{\arg\min}
\frac{1}{2}\|\mathbf{M}-\mathbf{M}^{k}\|^{2}_{F} + C\xi,\\
\mbox{s.t.} \thickspace  D_{\mathbf{M}}(\mathbf{p}, \mathbf{p}^{-})
- D_{\mathbf{M}}(\mathbf{p}, \mathbf{p}^{+})  \geq 1 -  \xi,
\thickspace \xi \geq  0,
\end{array}
\label{eq:online_optimization_function} \vspace{-0.22cm}
\end{equation}
where $\|\cdot\|_{F}$ denotes the Frobenius norm, $\xi$ is a slack
variable, and $C$ is a positive factor controlling
the trade-off between the smoothness term
$\frac{1}{2}\|\mathbf{M}-\mathbf{M}^{k}\|^{2}_{F}$ and the loss  term
$\xi$.
According to the passive-aggressive mechanism used
in~\cite{chechik2010large, crammer2006online}, we only update the
metric matrix $\mathbf{M}$ when $l_{\mathbf{M}}(\mathbf{p},
\mathbf{p}^{+}, \mathbf{p}^{-})>0$.

Subsequently, we derive an optimization
function with Lagrangian regularization:\vspace{-0.12cm}
\begin{equation}
\begin{array}{l}
\mathcal{L}(\mathbf{M}, \eta, \xi, \beta) =
\frac{1}{2}\|\mathbf{M}-\mathbf{M}^{k}\|^{2}_{F} + C\xi - \beta \xi\\
\hspace{0.9cm}+ \eta(1-\xi
+
D_{\mathbf{M}}(\mathbf{p},
\mathbf{p}^{+})-D_{\mathbf{M}}(\mathbf{p}, \mathbf{p}^{-})),
\end{array}
\label{eq:Lagrange_loss} \vspace{-0.12cm}
\end{equation}
where $\eta\geq 0$ and $\beta \geq 0$ are Lagrange multipliers. By taking the
derivative of $\mathcal{L}(\mathbf{M}, \eta, \xi, \beta)$
with respect to $\mathbf{M}$, we have the following: \vspace{-0.12cm}
\begin{equation}
\begin{array}{ll}
\frac{\partial\mathcal{L}(\mathbf{M}, \eta, \xi, \beta)}{\partial
\mathbf{M}} & = \mathbf{M}-\mathbf{M}^{k} +
\eta\frac{\partial
[D_{\mathbf{M}}(\mathbf{p},
\mathbf{p}^{+})-D_{\mathbf{M}}(\mathbf{p}, \mathbf{p}^{-})]
}
{\partial
\mathbf{M}}.\\
\end{array} \vspace{-0.12cm}
\end{equation}
Mathematically, $\frac{
\partial
[D_{\mathbf{M}}(\mathbf{p},
\mathbf{p}^{+})-D_{\mathbf{M}}(\mathbf{p}, \mathbf{p}^{-})]
}
{\partial \mathbf{M}}$ can be
formulated as: \vspace{-0.12cm}
\begin{equation}
\frac{\partial
[D_{\mathbf{M}}(\mathbf{p},
\mathbf{p}^{+})-D_{\mathbf{M}}(\mathbf{p}, \mathbf{p}^{-})]
}
{\partial \mathbf{M}}
=
\mathbf{a}_{+}\mathbf{a}_{+}^{T} - \mathbf{a}_{-}\mathbf{a}_{-}^{T},
\label{eq:triplet_update} \vspace{-0.12cm}
\end{equation}
where $\mathbf{a}_{+} = \mathbf{p} - \mathbf{p}^{+}$ and
$\mathbf{a}_{-} = \mathbf{p} - \mathbf{p}^{-}$. Therefore, the optimal
$\mathbf{M}^{k+1}$ is obtained
by setting $\frac{\partial\mathcal{L}(\mathbf{M}, \eta, \xi,
\beta)}{\partial \mathbf{M}}$ to zero. As a result,
the following relation holds: \vspace{-0.12cm}
\begin{equation}
\mathbf{M}^{k+1} = \mathbf{M}^{k} + \eta
(\mathbf{a}_{-}\mathbf{a}_{-}^{T} - \mathbf{a}_{+}\mathbf{a}_{+}^{T}).
\label{eq:eq:metric_update} \vspace{-0.12cm}
\end{equation}
Subsequently, we take the derivative of the
Lagrangian~\eqref{eq:Lagrange_loss} with respect to $\xi$ and set it
to zero: \vspace{-0.2cm}
\begin{equation}
\frac{\partial\mathcal{L}(\mathbf{M}, \eta, \xi, \beta)}{\partial
\xi} = C - \beta - \eta = 0.
\label{eq:lagrange_multiplier} \vspace{-0.2cm}
\end{equation}
Clearly, $\beta\geq 0$ leads to the fact that $\eta \leq C$.
For notational simplicity, $\mathbf{a}_{-}\mathbf{a}_{-}^{T} -
\mathbf{a}_{+}\mathbf{a}_{+}^{T}$ is abbreviated as $\mathbf{U}$
hereinafter.
By substituting Equs.~\eqref{eq:eq:metric_update}
and~\eqref{eq:lagrange_multiplier} into Equ.~\eqref{eq:Lagrange_loss} with $\mathbf{M} = \mathbf{M}^{k+1}$,  we have: \vspace{-0.2cm}
\begin{equation}
\mathcal{L}(\eta) = \frac{1}{2}\eta^{2}\|\mathbf{U}\|_{F}^{2} +
\eta(1+D_{\mathbf{M}^{k+1}}(\mathbf{p},
\mathbf{p}^{+})-D_{\mathbf{M}^{k+1}}(\mathbf{p}, \mathbf{p}^{-})),
\label{eq:Lagrange_eta} \vspace{-0.12cm}
\end{equation}
where $D_{\mathbf{M}^{k+1}}(\mathbf{p}, $ $ \mathbf{p}^{+}) =
\mathbf{a}_{+}^{T } $ $ (\mathbf{M}^{k}+\eta \mathbf{U})\mathbf{a}_{+}$ and
$D_{\mathbf{M}^{k+1}}(\mathbf{p}, $ $ \mathbf{p}^{-}) =
\mathbf{a}_{-}^{T}(\mathbf{M}^{k}+\eta \mathbf{U})\mathbf{a}_{-}$.
As a result, $\mathcal{L}(\eta)$ can be reformulated as: \vspace{-0.12cm}
\begin{equation}
\mathcal{L}(\eta) = \lambda_{2}\eta^{2} + \lambda_{1}\eta +
\lambda_{0}, \vspace{-0.016cm}
\end{equation}
where $\lambda_{2}= \frac{1}{2}\|\mathbf{U}\|_{F}^{2} +
\mathbf{a}_{+}^{T}\mathbf{U}\mathbf{a}_{+} -
\mathbf{a}_{-}^{T}\mathbf{U}\mathbf{a}_{-}$, $\lambda_{1} = 1 +
\mathbf{a}_{+}^{T}\mathbf{M}^{k}\mathbf{a}_{+} -
\mathbf{a}_{-}^{T}\mathbf{M}^{k}\mathbf{a}_{-}$,
and $\lambda_{0} = 0$. To obtain the optimal $\eta$, we need to
differentiate $\mathcal{L}(\eta)$ with respect to $\eta$ and set it to
zero: \vspace{-0.12cm}
\begin{equation}
\hspace{-0.05cm}
\begin{array}{l}
\frac{\partial \mathcal{L}(\eta)}{\partial \eta} =
\eta(\|\mathbf{U}\|_{F}^{2} +
2\mathbf{a}_{+}^{T}\mathbf{U}\mathbf{a}_{+} -
2\mathbf{a}_{-}^{T}\mathbf{U}\mathbf{a}_{-}) \\
      \hspace{2.1cm} + (1 +
\mathbf{a}_{+}^{T}\mathbf{M}^{k}\mathbf{a}_{+} -
\mathbf{a}_{-}^{T}\mathbf{M}^{k}\mathbf{a}_{-}) = 0.\\
\end{array} \vspace{-0.2cm} \hspace{-0.21cm}
\end{equation}
As a result, the following relation holds: \vspace{-0.2cm}
\begin{equation}
\eta = -\frac{1 + \mathbf{a}_{+}^{T}\mathbf{M}^{k}\mathbf{a}_{+} -
\mathbf{a}_{-}^{T}\mathbf{M}^{k}\mathbf{a}_{-}}{\|\mathbf{U}\|_{F}^{2} +
2\mathbf{a}_{+}^{T}\mathbf{U}\mathbf{a}_{+} -
2\mathbf{a}_{-}^{T}\mathbf{U}\mathbf{a}_{-}}.
\label{eq:eta_update} \vspace{-0.12cm}
\end{equation}
Due to the constraint of $0 \leq \eta\leq C$, $\eta$ should take the following
value: \vspace{-0.12cm}
\begin{equation}
\hspace{-0.0cm}
\eta \hspace{-0.08cm}= \hspace{-0.08cm}\min\left\{C, \max \left\{0,\frac{1 + \mathbf{a}_{+}^{T}\mathbf{M}^{k}\mathbf{a}_{+} -
\mathbf{a}_{-}^{T}\mathbf{M}^{k}\mathbf{a}_{-}}{2\mathbf{a}_{-}^{T}\mathbf{U}\mathbf{a}_{-} \hspace{-0.08cm}- \hspace{-0.08cm}
2\mathbf{a}_{+}^{T}\mathbf{U}\mathbf{a}_{+} \hspace{-0.08cm}- \hspace{-0.08cm}\|\mathbf{U}\|_{F}^{2}}\right\}\hspace{-0.1cm}\right\}
\hspace{-0.9cm}
\label{eq:eta_final} \vspace{-0.2cm}
\end{equation}
The complete procedure of online distance metric learning is
summarized in Algorithm~\ref{alg:online_metric_learning}.

\begin{algorithm}[t]
\begin{minipage}[ctb]{8.5cm}
\footnotesize
\caption{Time-weighted reservoir sampling}
\label{alg:Weighted_reserior_sampling}
\KwIn
{
 \hspace{-0.2cm} Current buffers $\mathcal{B}_{f}$ and  $\mathcal{B}_{b}$ together with their corresponding keys, a new
training sample $\mathbf{p}_{}$, maximum buffer size $\Omega$,
time-weighted factor $q$.
}
\KwOut
{
 \hspace{-0.1cm}Updated buffers $\mathcal{B}_{f}$ and  $\mathcal{B}_{b}$ together with their corresponding keys. \vspace{-0.39cm}
}
\begin{enumerate}\itemsep=-1.6pt
\item Obtain the samples $\mathbf{p}^{\ast}_{f}\in \mathcal{B}_{f}$ and $\mathbf{p}^{\ast}_{b} \in \mathcal{B}_{b}$ with the
smallest \mbox{\hspace{0.08cm}keys $k^{\ast}_{f}$ and $k^{\ast}_{b}$ from $\mathcal{B}_{f}$ and $\mathcal{B}_{b}$, respectively}.
  \item Compute the time-related weight $w = q^{\mathbb{I}_{}}$
with $\mathbb{I}_{}$ being the corresponding frame index  number of
$\mathbf{p}_{}$. \vspace{-0.05cm}
  \item Calculate a key $k_{}=u_{}^{1/w_{}}$ where $u_{}\sim rand(0,1)$.
  \item
  \textbf{Case:} $\mathbf{p}_{}\in$ foreground\\
  \mbox{\hspace{0.6cm}\textbf{if} $|\mathcal{B}_{f}|<\Omega$ \textbf{then}}  \vspace{-0.55cm}\\
         \begin{itemize}\itemindent=16pt
         \item $\mathcal{B}_{f}=\mathcal{B}_{f}\bigcup \{\mathbf{p}_{}\}$.
          \vspace{-0.28cm}
         \end{itemize}
  \mbox{\hspace{0.6cm}\textbf{else}} \vspace{-0.26cm}
         \begin{itemize} \itemindent=16pt
         \item $\mathbf{p}^{\ast}_{f}$ is replaced with $\mathbf{p}_{}$ if \mbox{\hspace{0.08cm}$k_{}>k^{\ast}_{f}$}.
           \vspace{-0.2cm}
         \end{itemize}
  \mbox{\hspace{0.6cm}\textbf{endif}}\vspace{-0.0cm}\\
        \textbf{Case:} $\mathbf{p}_{}\in$ background\\
  \mbox{\hspace{0.6cm}\textbf{if} $|\mathcal{B}_{b}|<\Omega$ \textbf{then}}  \vspace{-0.55cm}\\
         \begin{itemize}\itemindent=16pt
         \item $\mathcal{B}_{b}=\mathcal{B}_{b}\bigcup \{\mathbf{p}_{}\}$.
          \vspace{-0.28cm}
         \end{itemize}
  \mbox{\hspace{0.6cm}\textbf{else}} \vspace{-0.26cm}
         \begin{itemize} \itemindent=16pt
         \item $\mathbf{p}^{\ast}_{b}$ is replaced with $\mathbf{p}_{}$ if $k_{}>k^{\ast}_{b}$.
          \vspace{-0.2cm}
         \end{itemize}
  \mbox{\hspace{0.6cm}\textbf{endif}}\vspace{-0.3cm}\\
   \item Return $\mathcal{B}_{f}$ and  $\mathcal{B}_{b}$ together with their corresponding keys.
\end{enumerate}
\end{minipage}
\end{algorithm}

\subsection{Time-weighted reservoir sampling}
\label{sec:twrs}

We compute a linear representation solution
(Equ.~\eqref{eq:regression_optimal_solution}) for two separate sample buffers consisting of
foreground and background basis samples. Ideally, the sample buffers should keep a balance
between sample diversity and  adaptability.
Motivated by this, reservoir sampling~\cite{vitter1985random,
zhaoICML2011, Kolonko04sequentialreservoir, efraimidis2006weighted} is proposed for sequential
random sampling. In principle, it aims to
randomly draw some samples from a large population of samples that
come in a sequential manner. A classical version of
reservoir sampling is able to effectively simulate the process of
uniform random sampling~\cite{vitter1985random, zhaoICML2011}.
However, it is inappropriate for visual
tracking because the samples used in visual tracking are dynamically
distributed as time progresses. Usually, the samples occurring
recently have
a greater influence on the current tracking process than those
appearing a long time ago. Therefore, larger weights should be assigned
to the recently added samples
while smaller weights should be attached with the old samples. Inspired
by~\cite{Kolonko04sequentialreservoir,efraimidis2006weighted}, we design a time-weighted reservoir
sampling (TWRS)
method for randomly drawing the samples according to their
time-varying properties, as listed in
Algorithm~\ref{alg:Weighted_reserior_sampling}.
The designed TWRS method is capable of effectively maintaining
the sample buffers for online metric learning in Sec.~\ref{sec:oml}.

\begin{algorithm}[t]
\caption{Metric-weighted linear representation based visual tracking with
time-weighted reservoir sampling}
\label{alg:Framwork}
\footnotesize{
\textbf{Input:} Frame $t$, previous object state
$\textbf{Z}_{t-1}^{\ast}$, previous distance metric matrix
 $\mathbf{M}_{t-1}$, foreground buffer $\mathcal{B}_{f}$ with its basis samples $\mathbf{P}_{f}$,
background  buffer $\mathcal{B}_{b}$ with its basis samples $\mathbf{P}_{b}$, number of particles $\mathcal{K}$. \\
\textbf{Output:} Current object state $\textbf{Z}_{t}^{\ast}$, updated
metric matrix $\mathbf{M}_{t}$, updated $\mathcal{B}_{f}$ and
$\mathcal{B}_{b}$.
\begin{algorithmic}[1]
\STATE Sample a number of candidate object states
$\{\textbf{Z}_{t}^{k}\}_{k=1}^{\mathcal{K}}$ using the particle filters
(i.e., Gaussian dynamical model used in \cite{Limy-Ross17}).
\STATE Crop out the corresponding image regions of
$\{\textbf{Z}_{t}^{k}\}_{k=1}^{\mathcal{K}}$.
\STATE Extract the corresponding HOG feature set
$\{\mathbf{y}_{k}\}_{k=1}^{\mathcal{K}}$.
\STATE Perform the metric-weighted optimization in
Equ.~\eqref{eq:linear_regression_metric} with
$\min_{\mathbf{x}_{f}}  g(\mathbf{x}_{f}; \mathbf{M}_{t-1},
\mathbf{P}_{f}, \mathbf{y}_{k})$
and $\min_{\mathbf{x}_{b}}  g(\mathbf{x}_{b}; \mathbf{M}_{t-1},
\mathbf{P}_{b}, \mathbf{y}_{k})$.
\STATE Determine the optimal object state $\textbf{Z}_{t}^{\ast}$ by
the MAP (maximum a posterior) estimation in the particle filters,
where the observation model is defined in
Equ.~\eqref{eq:particle_liki_model} such that
$p\left(\mathbf{y}_{k}|\mathbf{Z}_{t}^{k}\right)  \propto
\mathcal{S}(\mathbf{y}_{k})$.
\STATE Collect new foreground and background samples
$\mathcal{P}_{f}\bigcup \mathcal{P}_{b}$ according to the spatial
distance-based mechanism of training sample selection.
\STATE Carry out time-weighted reservoir sampling in
Algorithm~\ref{alg:Weighted_reserior_sampling} to iteratively update
$\mathcal{B}_{f}$ and $\mathcal{B}_{b}$ with new training samples from
$\mathcal{P}_{f}\bigcup \mathcal{P}_{b}$.
\STATE Perform the triplet sampling procedure (s.t. intra-class
relevance and inter-class irrelevance) in~\cite{chechik2010large} over
$\mathcal{B}_{f}\bigcup\mathcal{B}_{b}$ to generate a triplet set $\mathcal{Q} = \{(\mathbf{p}_{}, \mathbf{p}^{+},
\mathbf{p}^{-})\}$.
\STATE Run online metric learning in
Algorithm~\ref{alg:online_metric_learning} to update
$\mathbf{M}_{t-1}$ for each triplet in $\mathcal{Q}$, and finally
obtain $\mathbf{M}_{t}$. This step can be performed every few frames.
\STATE Return $\textbf{Z}_{t}^{\ast}$, $\mathbf{M}_{t}$,
$\mathcal{B}_{f}$, and $\mathcal{B}_{b}$.
\end{algorithmic}
}
\vspace{-0.0cm}
\end{algorithm}

By integrating the above-mentioned three modules (i.e., metric-weighted linear representation, online metric learning,
and time-weighted reservoir sampling) into a particle filtering framework, we obtain a visual tracker whose
complete procedure is shown in
Algorithm~\ref{alg:Framwork}.

\section{Experiments}

{\bf Experimental setup}
In order to evaluate the proposed tracking algorithm, we conduct a set
of experiments on
thirteen challenging video sequences consisting of
8-bit grayscale images.
These video sequences are
captured from different scenes,
and contain a variety of object motion events (e.g., human walking and
car running).

The proposed tracking algorithm is implemented in Matlab
on a workstation
with an Intel Core 2 Duo 2.66GHz processor and 3.24G RAM.
The average running time of the proposed tracking algorithm is about
0.55 second per frame.
For the sake of computational efficiency, we simply consider the
object state information in 2D translation and
scaling in the particle filtering module, where
the corresponding variance parameters are set to (10, 10, 0.1).
The number of particles is set to 200.
For each particle, there is a corresponding image region
represented as a HOG feature descriptor (referred to \cite{Dalal-Triggs-CVPR2005} and efficiently computed by using integral histograms) with $3\times 3$ cells
(each cell is represented by a 9-dimensional histogram vector)
in the five spatial block-division modes (like~\cite{lixi-cvpr2008}),
resulting in
a 405-dimensional feature vector for the image region.
The number of triplets used for online metric learning
is chosen as 500.
The maximum buffer size $\Omega$ and time-weighted factor $q$ in
Algorithm~\ref{alg:Weighted_reserior_sampling}
is set as 300 and 1.6, respectively. The scaling factors $\gamma_{f}$
and $\gamma_{b}$ in Equ.~\eqref{eq:particle_liki_model} are
chosen as 1. The trade-off control factor $\rho$ in
Equ.~\eqref{eq:particle_liki_model} is set as 0.1.
Note that the aforementioned parameters are fixed throughout all the
experiments.

\begin{figure}[t]
\centering
\includegraphics[scale=0.36]{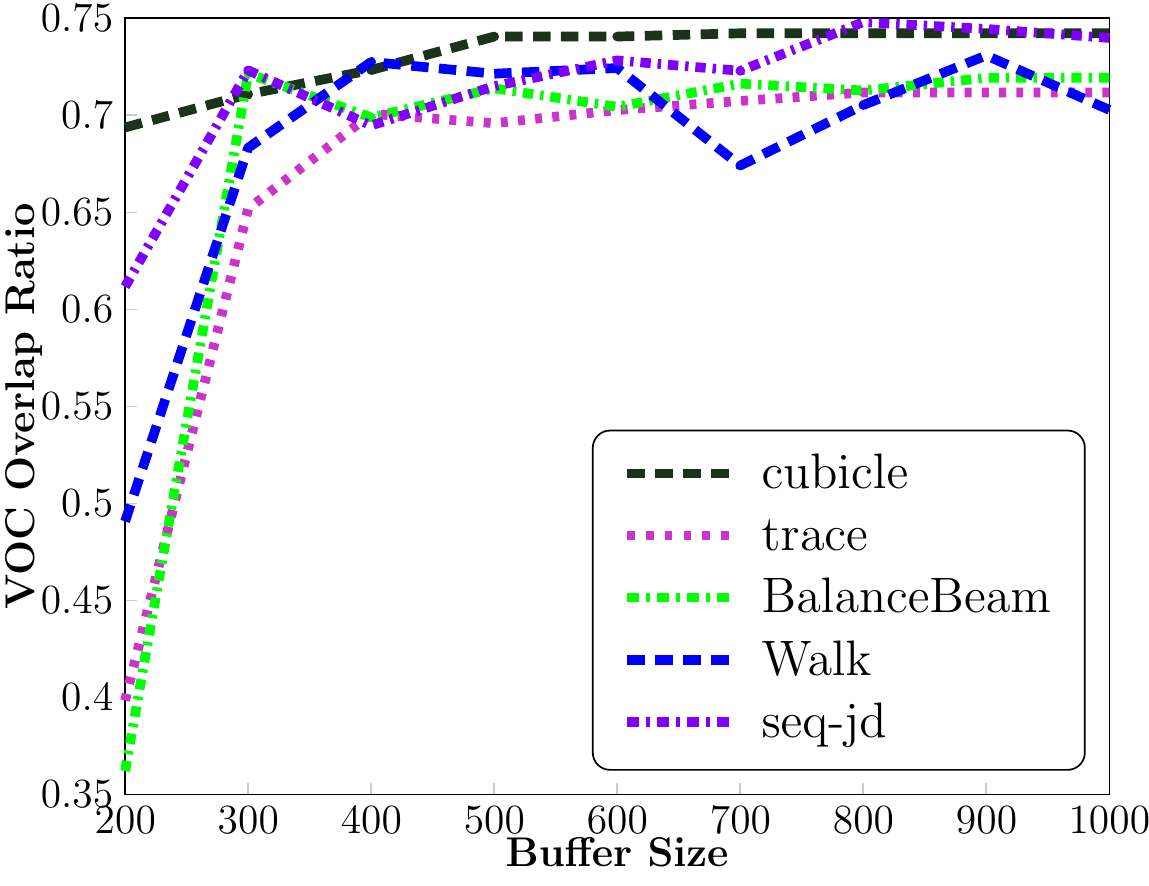}
\hspace{-0.2cm}
\includegraphics[scale=0.36]{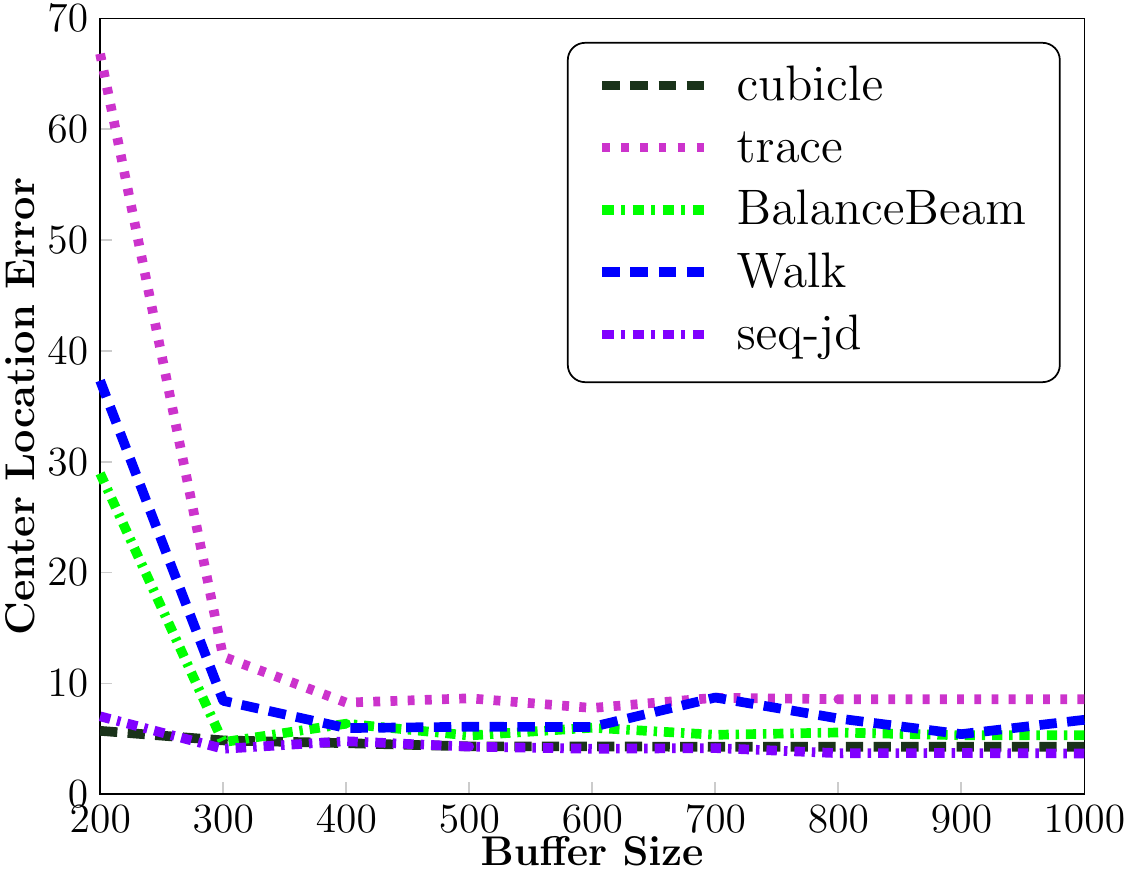}\\
\vspace{-0.16cm}
\vspace{-0.2cm}
 \caption{Quantitative evaluation of the proposed tracker using
different buffer sizes  on five video sequences (i.e., ``cubicle'',
``trace'', ``BalanceBeam'',
 ``Walk'', and ``seq-jd''). The left and right subfigures correspond
 to the tracking performance of the proposed tracking algorithm in
VOR and CLE, respectively. \vspace{-0.1cm}}
 \label{fig:buffersize}
\end{figure}

\begin{figure}[t]
\centering
\includegraphics[scale=0.36]{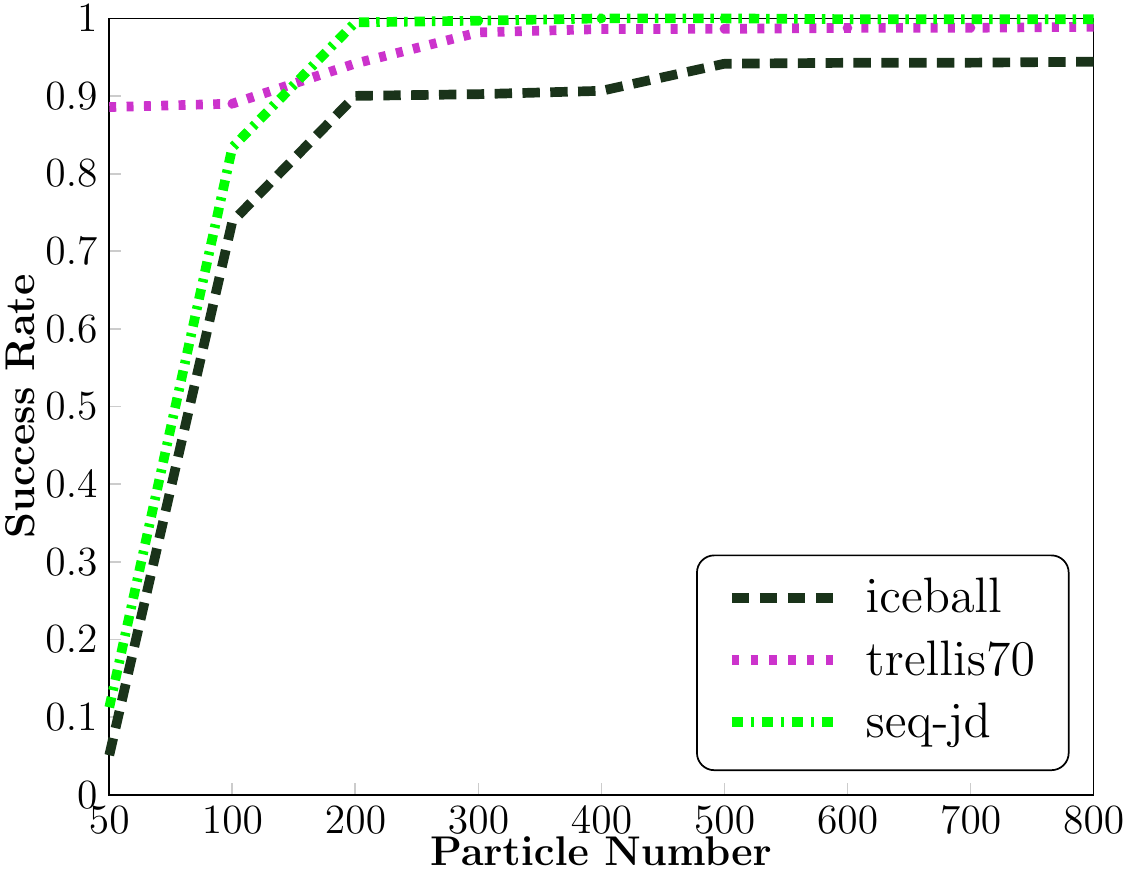}
\hspace{-0.2cm}
\includegraphics[scale=0.36]{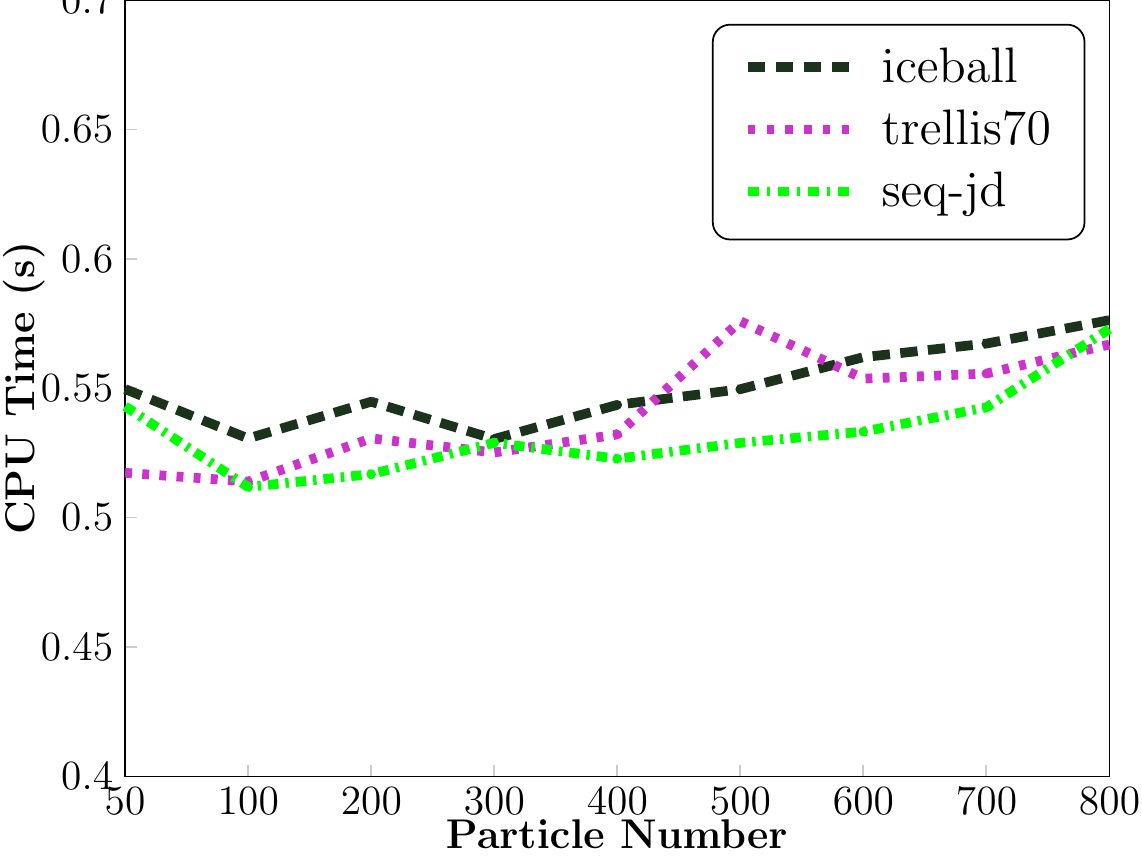}\\
\vspace{-0.13cm}
 \vspace{-0.2cm}
 \caption{Quantitative evaluation of the proposed tracker using
different particle numbers on three video sequences (i.e.,
``iceball'', ``trellis70'', and ``seq-jd''). The left and right
subfigures are associated with
 the tracking performance in average VOC success rate and tracking
duration for each frame, respectively. \vspace{-0.15cm}}
 \label{fig:particle_number}
\end{figure}

To demonstrate the effectiveness of the proposed tracking algorithm, we
compare it with other state-of-the-art trackers in both
qualitatively and quantitatively.
These trackers are referred to as
FragT (Fragment-based tracker \cite{Adam-Fragment-2006}),
MILT (multiple instance boosting-based
tracker~\cite{Babenko-Yang-Belongie-cvpr2009}), VTD (visual tracking
decomposition~\cite{Kwon-Lee-CVPR2010}), OAB (online
AdaBoost \cite{Grabner-Grabner-Bischof-BMVC2006}), IPCA (incremental
PCA~\cite{Limy-Ross17}), L1T ($\ell_{1}$ minimization
tracker~\cite{Meo-Ling-ICCV09}),
and DMLT (discriminative metric learning
tracker~\cite{wang2010discriminative}). In the experiments, some of
the aforementioned
trackers are implemented using their publicly available source code, including
FragT,
MILT,
VTD,
OAB,
IPCA,
and L1T.
For OAB, there are two different versions (namely, OAB1 and OAB5), which
are based on two different configurations (i.e., the search scale $r=1$ and $r=5$
as in \cite{Babenko-Yang-Belongie-cvpr2009}).
For quantitative performance comparison, two popular evaluation
criteria are introduced, namely,
center location error (CLE) and VOC overlap ratio (VOR)
between the predicted bounding
box $B_p$ and ground truth bounding box $B_{gt}$
such that ${\rm VOR} = \frac{area(B_p\bigcap B_{gt})}{area(B_p \bigcup B_{gt})}$.
If the
VOC overlap ratio
is larger than $0.5$, then it is considered  successful 
tracking.

{\bf Effect of different buffer sizes}
We aim to investigate the effect of using different buffer sizes for
visual tracking. Motivated by this,
a quantitative evaluation of the proposed tracking algorithm
is performed in nine different cases of
buffer size. Meanwhile, we compute the average
CLE and VOR for each video sequence in each case of buffer size.
Fig.~\ref{fig:buffersize} shows the quantitative CLE and
VOR performance on five video sequences.
It is clear that the average CLE (VOR) decreases (increases) as the
buffer size increases, and
plateaus with approximately more than 300 samples.

{\bf Evaluation of different particle numbers}
In general, more particle numbers
enable visual trackers to locate the object more accurately, but lead
to a higher computational cost.
Thus, it is crucial for visual trackers to keep a good balance between
accuracy and efficiency using a
moderate number of particles. Motivated by this, we examine the
tracking performance of the proposed
tracking algorithm with respect to different particle numbers.
The left part of Fig.~\ref{fig:particle_number} shows the average VOC success rates
(i.e., $\frac{\#\mbox{success frames}}{\# \mbox{total frames}}$)
of the proposed tracking algorithm on three video sequences.
From the left part of Fig.~\ref{fig:particle_number}, we can see that the success
rate  rapidly grows with the increase of
particle number and finally converges.
The right part of Fig.~\ref{fig:particle_number} displays the average
CPU time (spent by the proposed tracking algorithm in each frame) with
different particle numbers.
It is observed from the right part of Fig.~\ref{fig:particle_number} that the average
CPU time slowly increase.

\begin{figure}[t]
\centering
\includegraphics[scale=0.36]{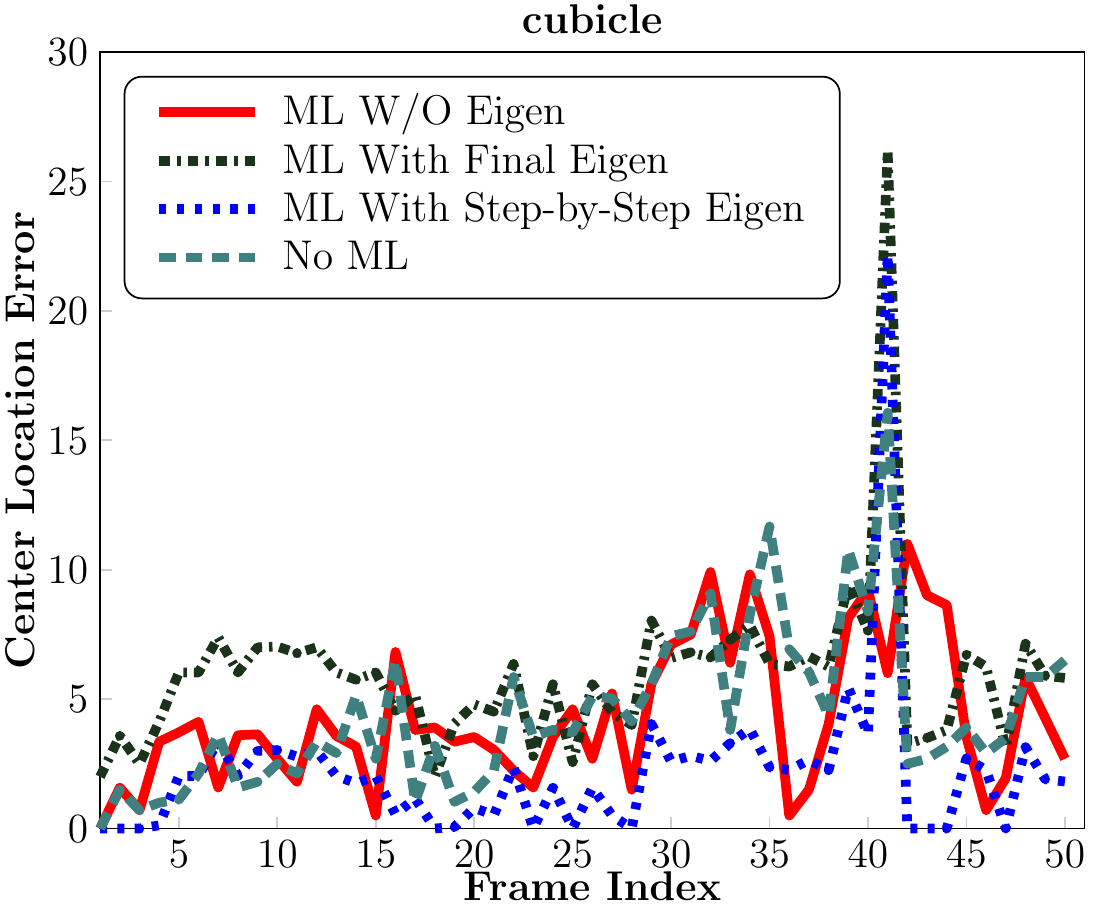}
\hspace{-0.13cm}
\includegraphics[scale=0.36]{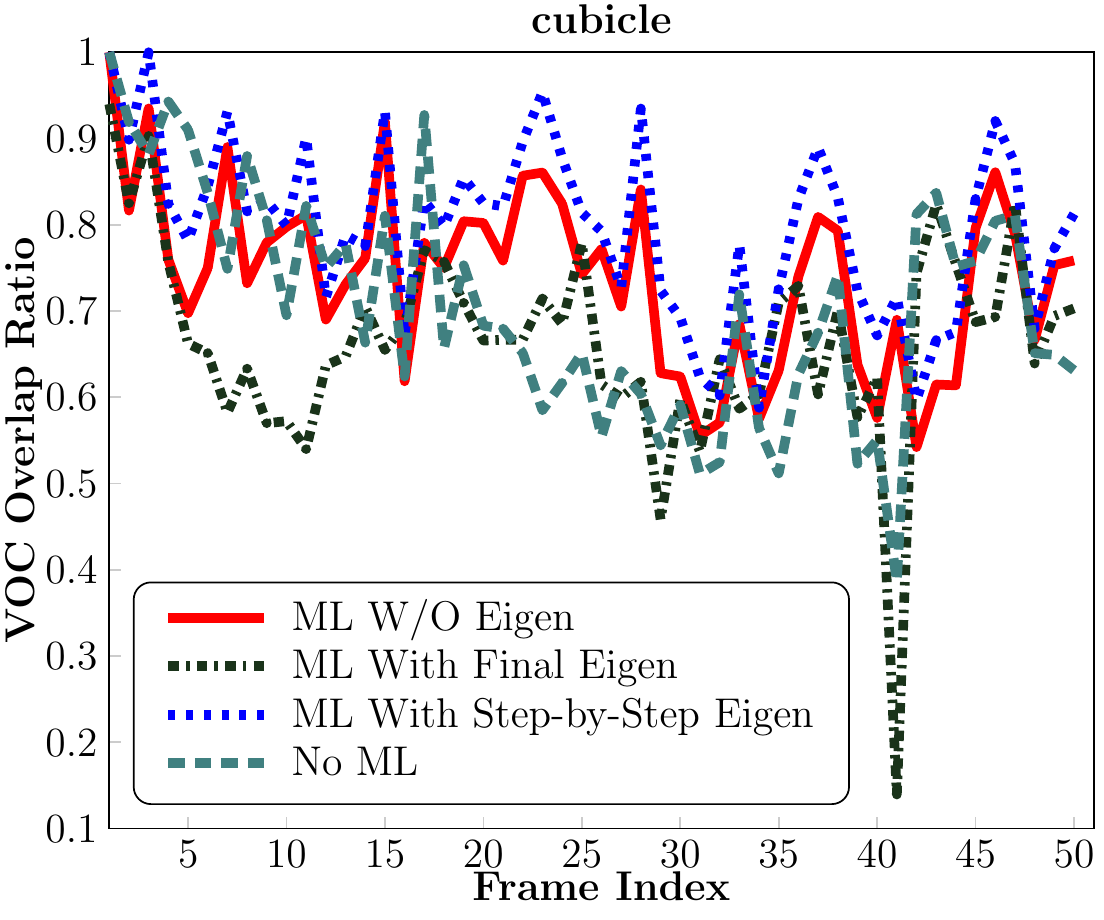}\\
\includegraphics[scale=0.36]{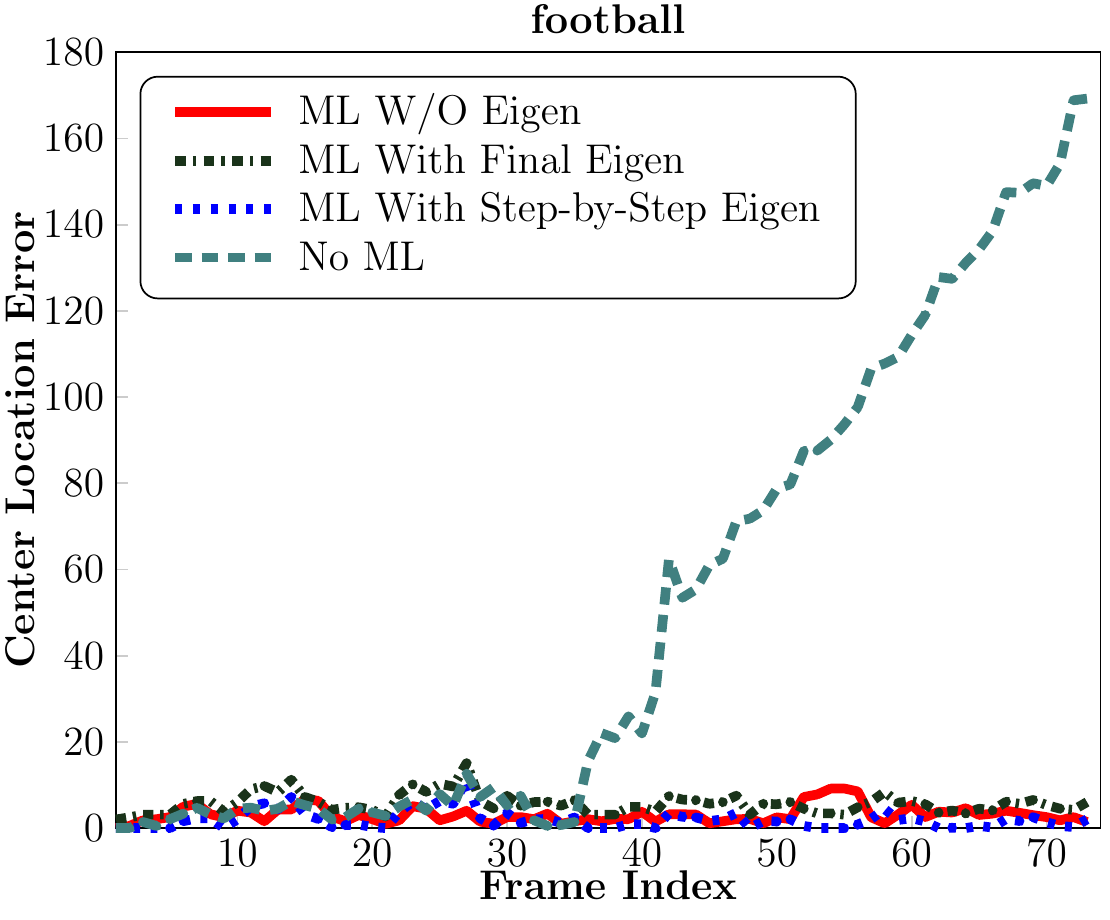}
\hspace{-0.13cm}
\includegraphics[scale=0.36]{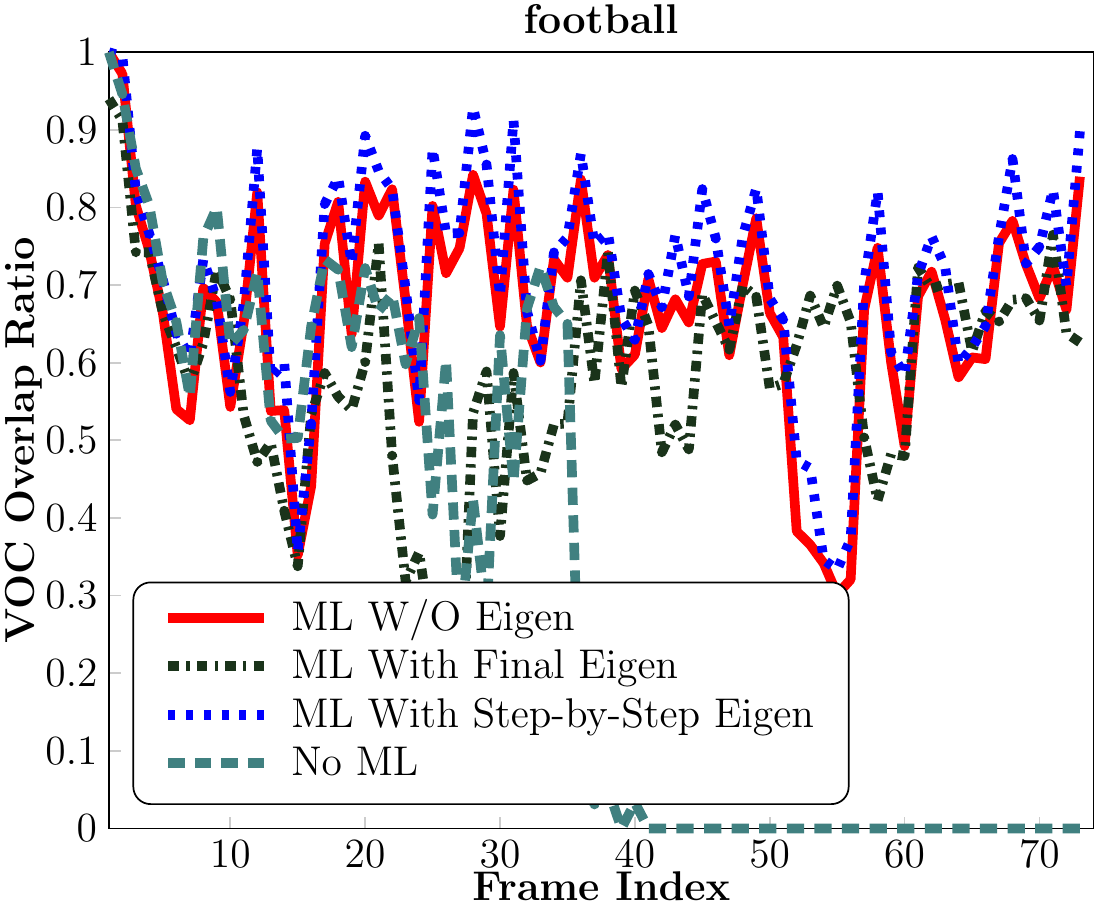}\\
\vspace{-0.16cm}
 \vspace{-0.2cm}
 \caption{Quantitative evaluation of the proposed tracker with/without
metric learning on two video sequences. The top two subfigures are
associated with
 the tracking performance in CLE and VOR on the ``cubicle'' video
sequence, respectively; the bottom two subfigures correspond to
 the tracking performance in CLE and VOR on the ``football'' video
sequence, respectively.
 \vspace{-0.16cm}}
 \label{fig:metric_non_metric}
\end{figure}

{\bf Performance with and without metric learning}
Metric learning is able to improve the intra-class compactness and
inter-class separability of
samples.
In metric learning, three types of learning mechanisms can be used, including
no eigendecomposition, step-by-step eigendecomposition, and final
eigendecomposition~\cite{chechik2010large}.
To justify the effect of different metric learning mechanisms, we
design several experiments on four video sequences.
Fig.~\ref{fig:metric_non_metric} shows the corresponding experimental
results of different metric learning mechanisms
in both CLE and VOR on two of the four video sequences (note that the results for
the other two video sequences can be found in the supplementary file).
Tab.~\ref{Tab:metric_success_rate} reports the
average success rates of different metric learning mechanisms on the four
video sequences.
From Fig.~\ref{fig:metric_non_metric} and
Tab.~\ref{Tab:metric_success_rate}, we can see that the performance of
metric learning is  better than that of no
metric learning. In addition, the performance of metric learning with
no eigendecomposition is close
to that of metric learning with step-by-step eigendecomposition, and
better than that of
metric learning with final eigendecomposition. Therefore, the obtained
results are consistent with those
in~\cite{chechik2010large}. Besides, metric learning with step-by-step
eigendecomposition
is much slower than that with no eigendecomposition which is adopted
by the proposed
tracking algorithm.

\begin{table}[t]
\scalebox{0.83}
{
\begin{tabular}{c||c|c|c|c}
\hline \scriptsize
& \makebox[0.9cm]{cubicle}   &  \makebox[0.9cm]{football} &
\makebox[0.9cm]{iceball} & \makebox[0.9cm]{trellis70} \\\hline
ML w/o eigen          &   \bf  0.98  &   0.88      &  0.93       & 0.98\\
ML with final eigen       &    0.94      &  0.74       &    0.90     &  0.94\\
ML with step-by-step eigen&  \bf   0.98  & \bf  0.90   &  \bf 0.95
& \bf 0.99\\
No metric learning                     &   0.86       &   0.36      &    0.88      &  0.91\\
\hline
\end{tabular}
}\vspace{-0.27cm}
\caption{Quantitative evaluation of the proposed tracker with/without
metric learning on four video sequences
The table reports their average success rates for each video sequence.
 \vspace{-0.1cm}}
\label{Tab:metric_success_rate}
\end{table}

\begin{figure}[t]
\centering
\includegraphics[scale=0.36]{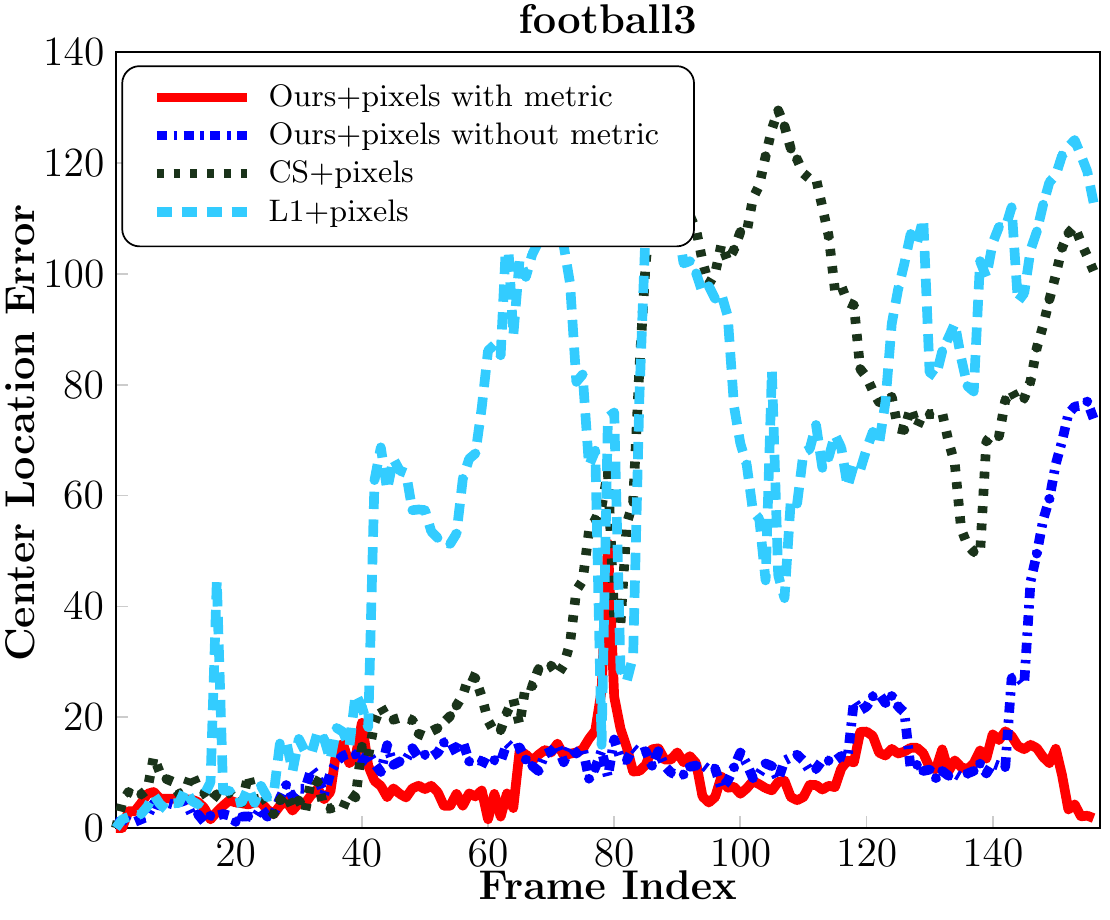}
\hspace{-0.13cm}
\includegraphics[scale=0.36]{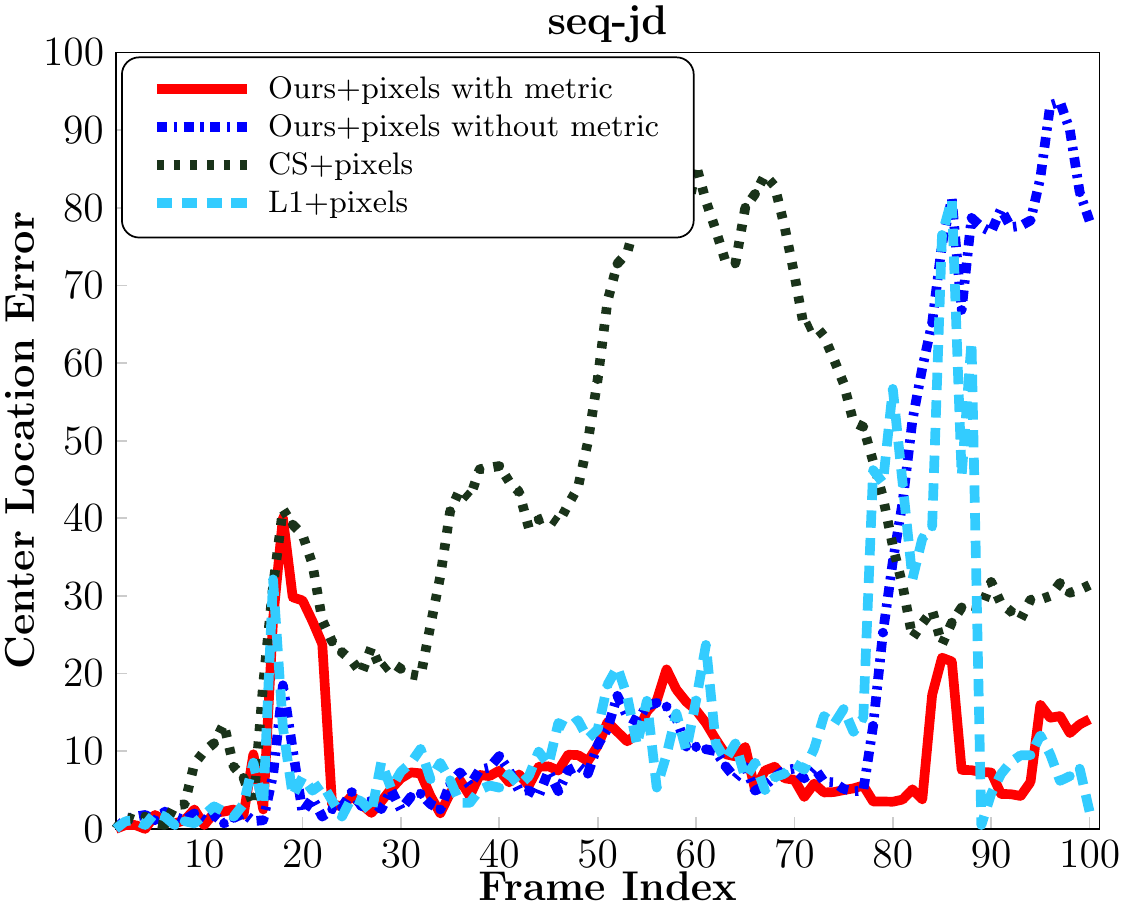}\\
\includegraphics[scale=0.36]{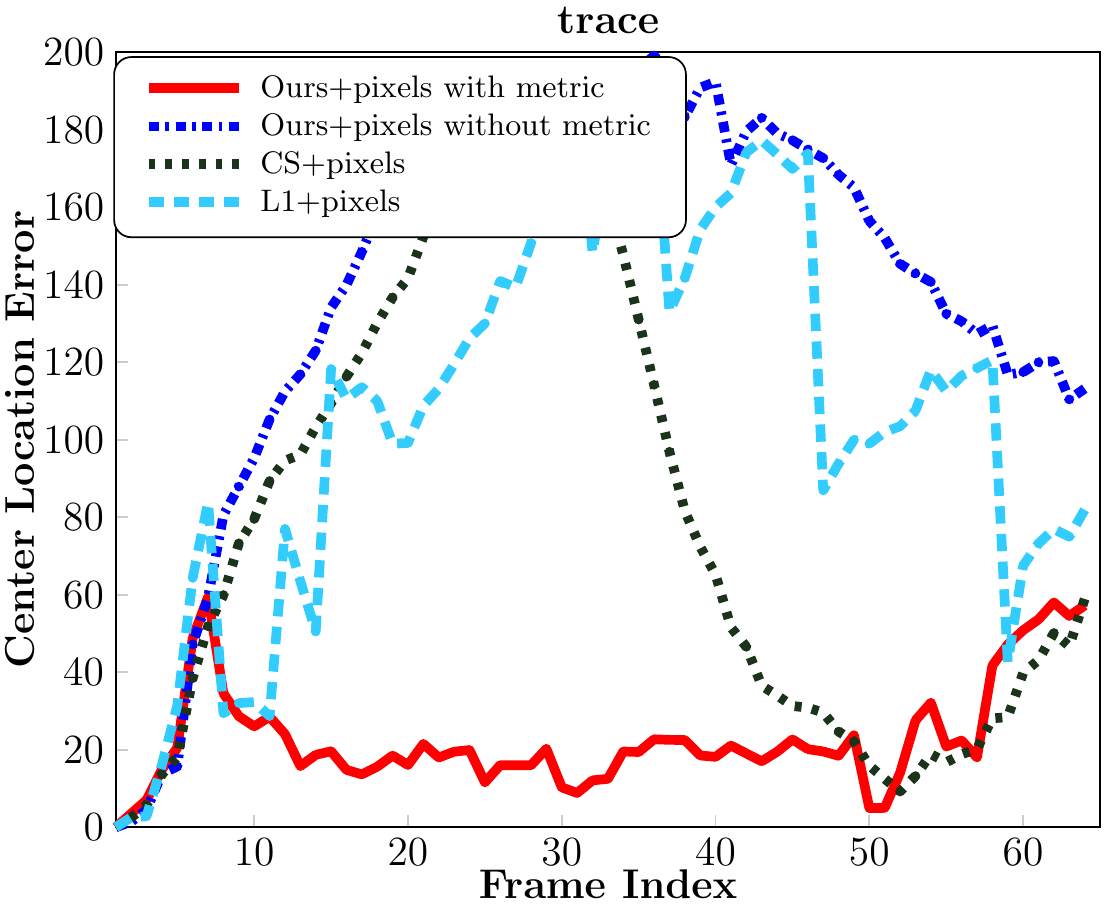}
\hspace{-0.13cm}
\includegraphics[scale=0.36]{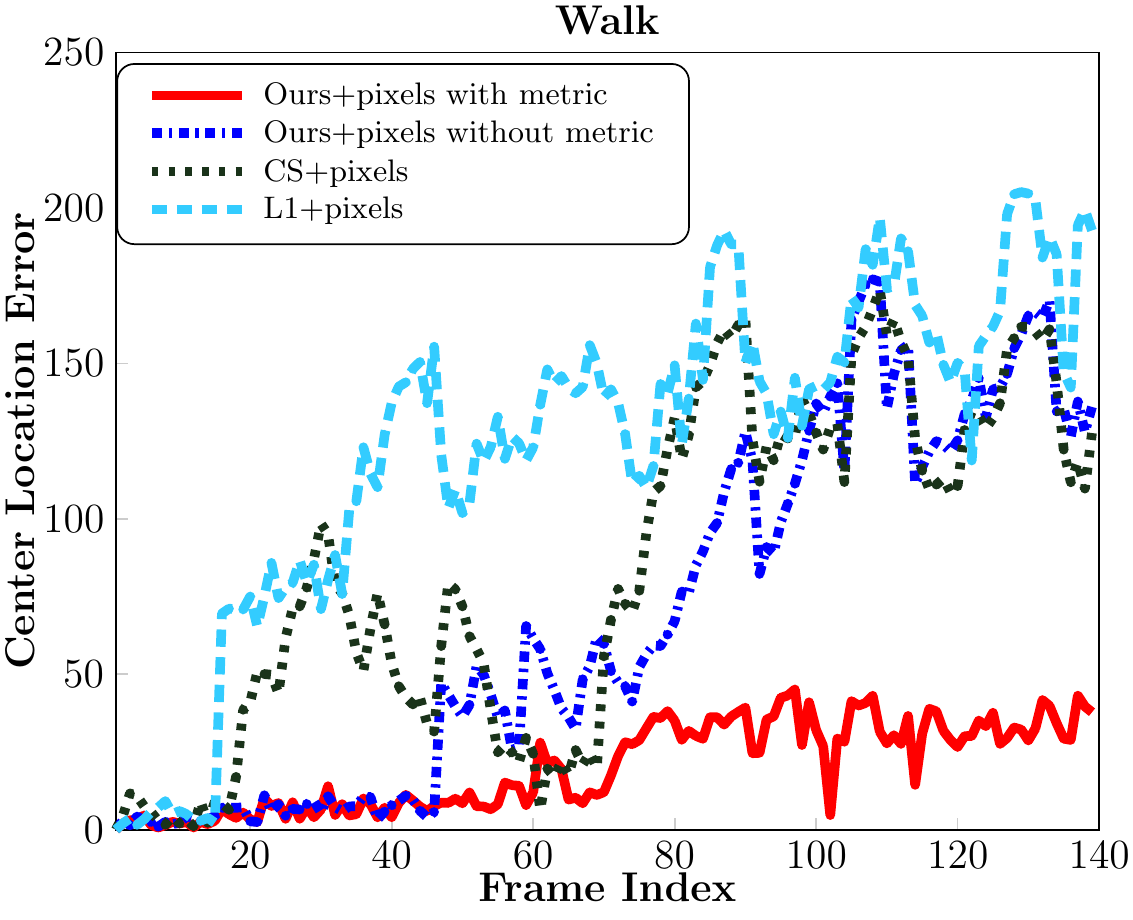}\\
\vspace{-0.16cm}
\vspace{-0.2cm}
 \caption{Quantitative comparison of different linear representation
methods in CLE on four video sequences (i.e.,
 ``football3'', ``seq-jd'', ``trace'', and ``Walk'').
 \vspace{-0.15cm}}
 \label{fig:regression}
\end{figure}

{\bf Comparison of different linear representations}
The objective of this task is to evaluate the performance of four types of
linear representations including our linear
representation with metric learning,
our linear representation without
metric learning,
compressive sensing linear representation~\cite{Li-Shen-Shi-cvpr2011},
and $\ell_{1}$-regularized linear
representation~\cite{Meo-Ling-ICCV09}.
For a fair comparison, we utilize the raw pixel features which are the same as
\cite{Li-Shen-Shi-cvpr2011,Meo-Ling-ICCV09}.
Fig.~\ref{fig:regression} shows the performance of these four linear
representation methods
in CLE on four video sequences. Clearly, our linear
representation with metric learning consistently achieves
lower CLE performance in most frames than the three other linear representations.

\begin{figure*}[t]
\centering
\includegraphics[scale=0.3571]{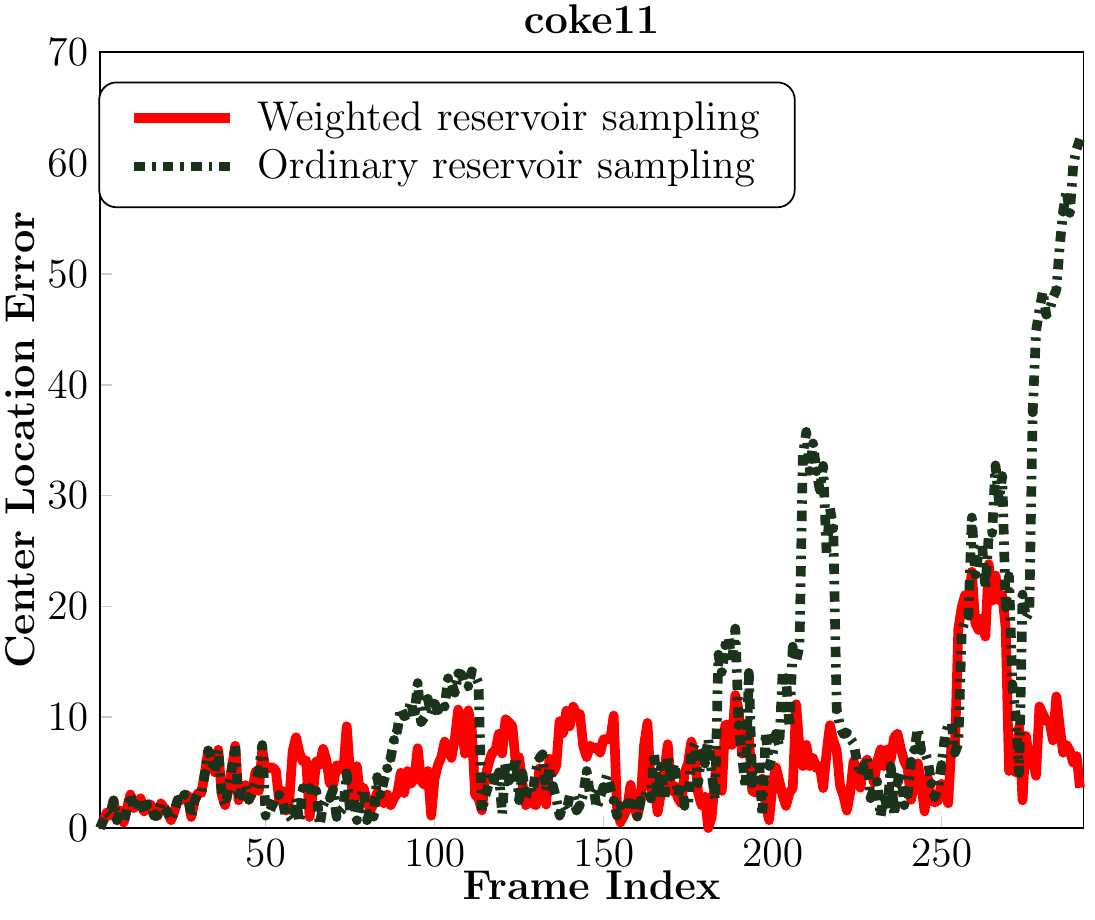}
\includegraphics[scale=0.3571]{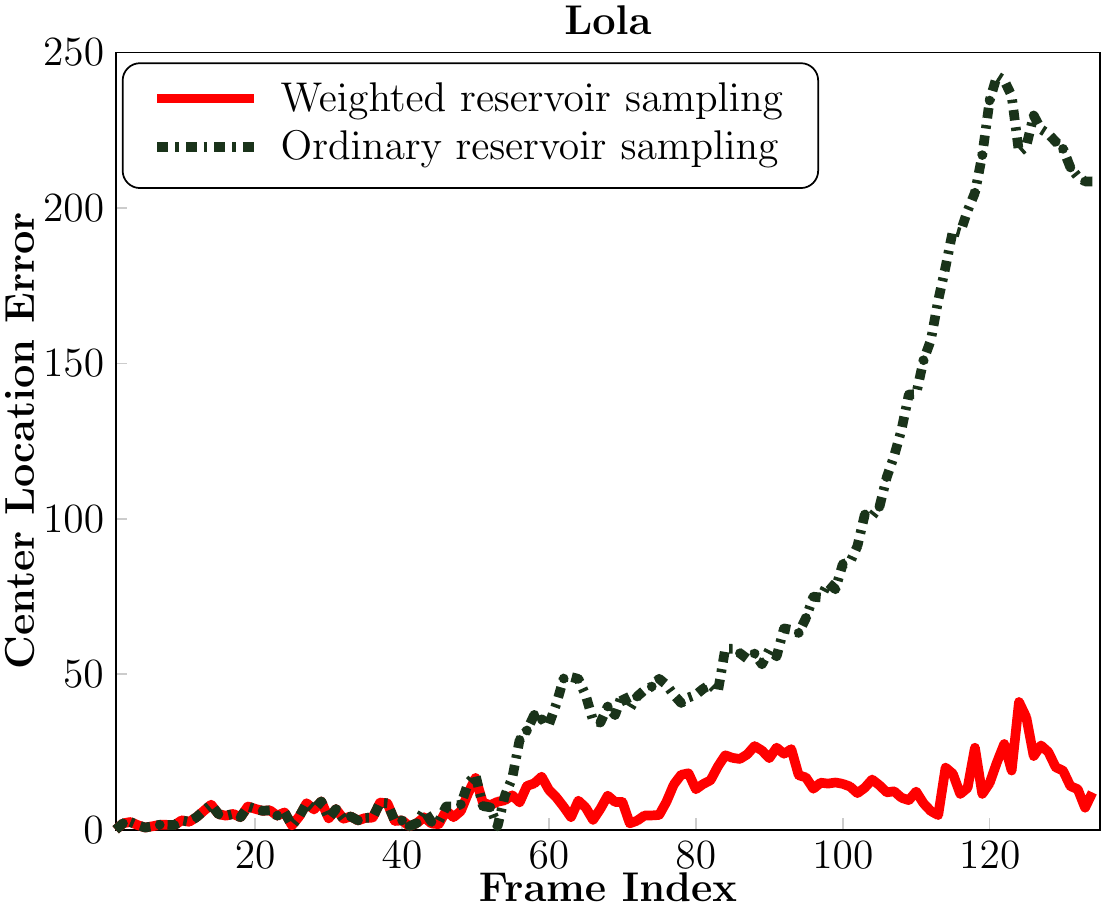}
\includegraphics[scale=0.3571]{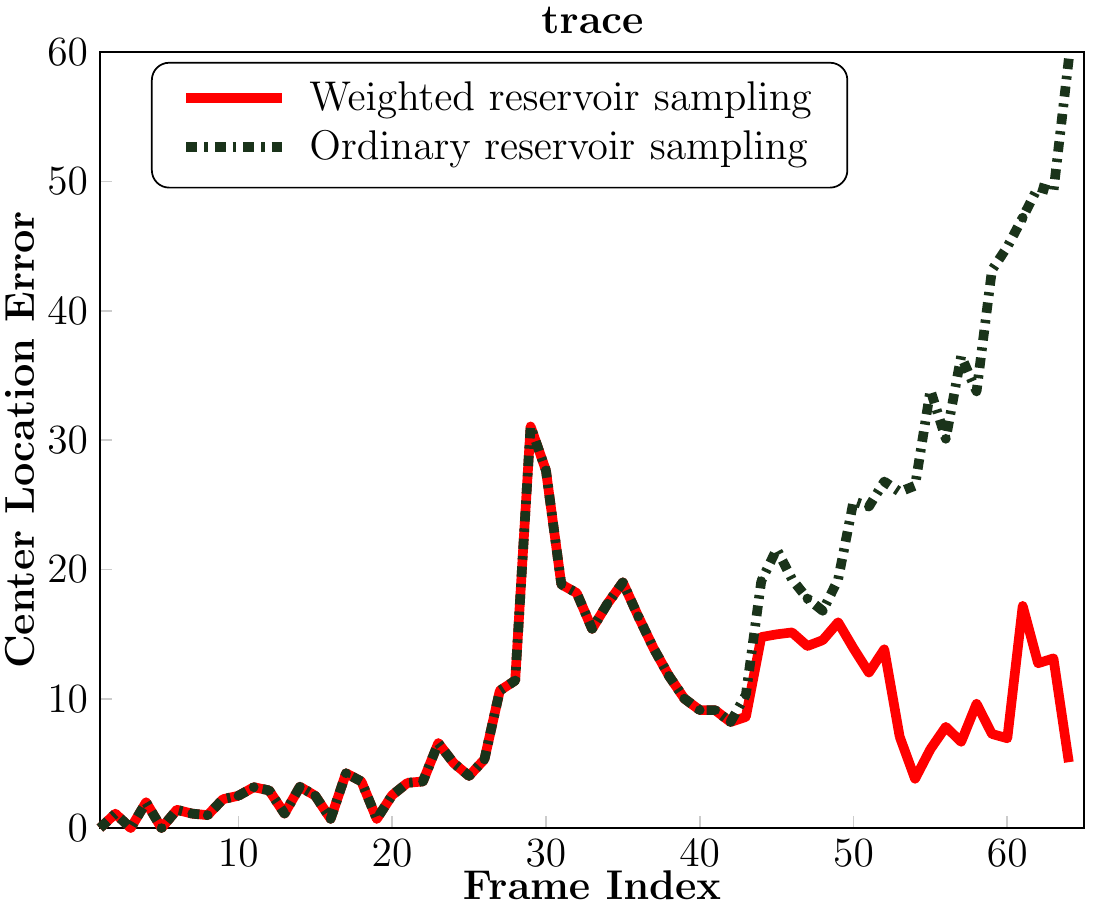}
\includegraphics[scale=0.3571]{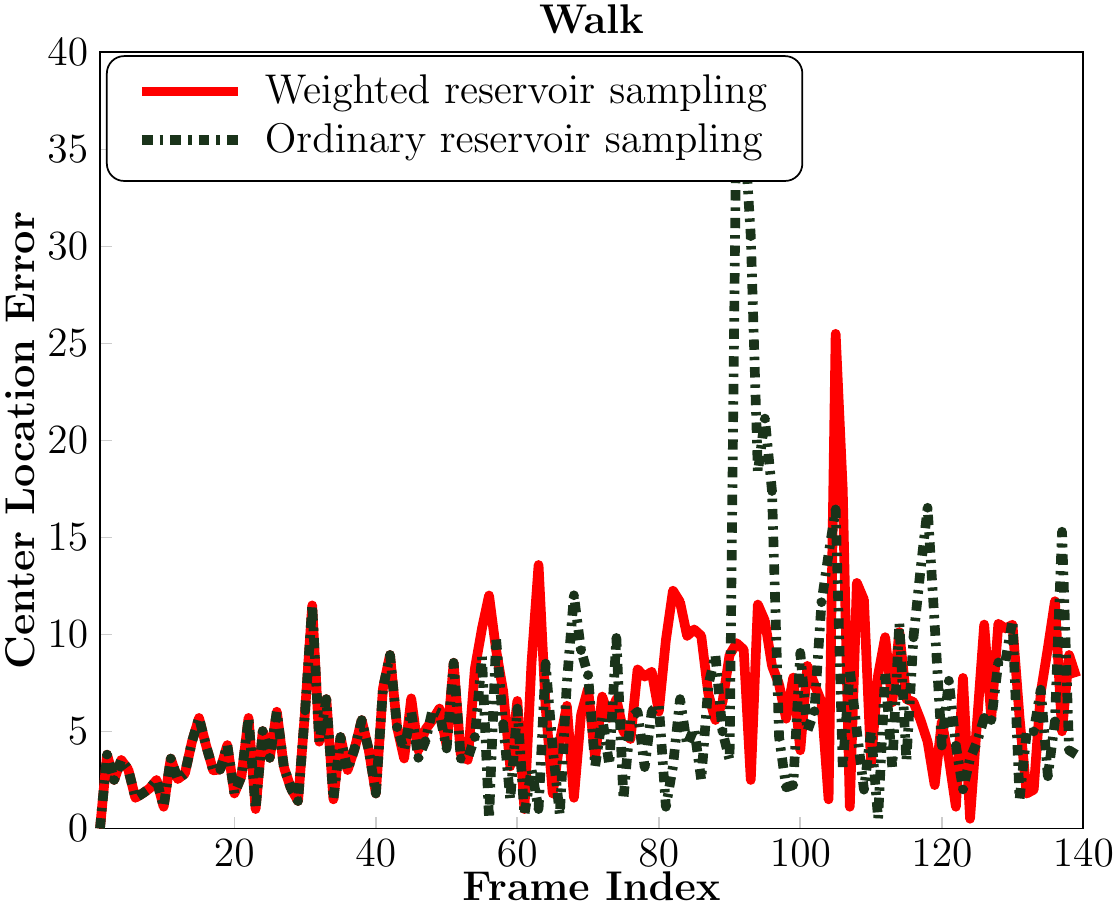}\\
\vspace{-0.3cm}
\caption{Quantitative comparison of different sampling methods in
CLE on four video sequences (i.e.,
 ``coke11'', ``Lola'', ``trace'', and ``Walk'').
Before exceeding the buffer size limit (approximately occurring between frame 40 and
frame 50), the performances
 of different sampling methods are identical.
 \vspace{-0.1cm}}
  \label{fig:sampling}
\end{figure*}

{\bf Evaluation of different sampling methods}
Reservoir sampling~\cite{vitter1985random} addresses the problem of
randomly drawing the uniformly distributed samples
in a sequential manner. Following the work of~\cite{vitter1985random},
a weighted version of reservoir sampling
is proposed in~\cite{efraimidis2006weighted}, which assign different
weights to the samples
occurring at different time points. Based on this weighed reservoir
sampling method,
the proposed tracking algorithm is capable of adaptively updating the
sample buffer as
tracking proceeds. Here, we aim to examine the performance of the two
sampling methods.
Fig.~\ref{fig:sampling} shows the experimental results of the two
sampling methods in CLE on four video sequences (note that the VOR results for these
four video sequences can be found in the supplementary file).
From Fig.~\ref{fig:sampling}, we can see that weighted reservoir
sampling performs better than
ordinary reservoir sampling.

\begin{figure*}[t]
\centering
\includegraphics[scale=0.3571]{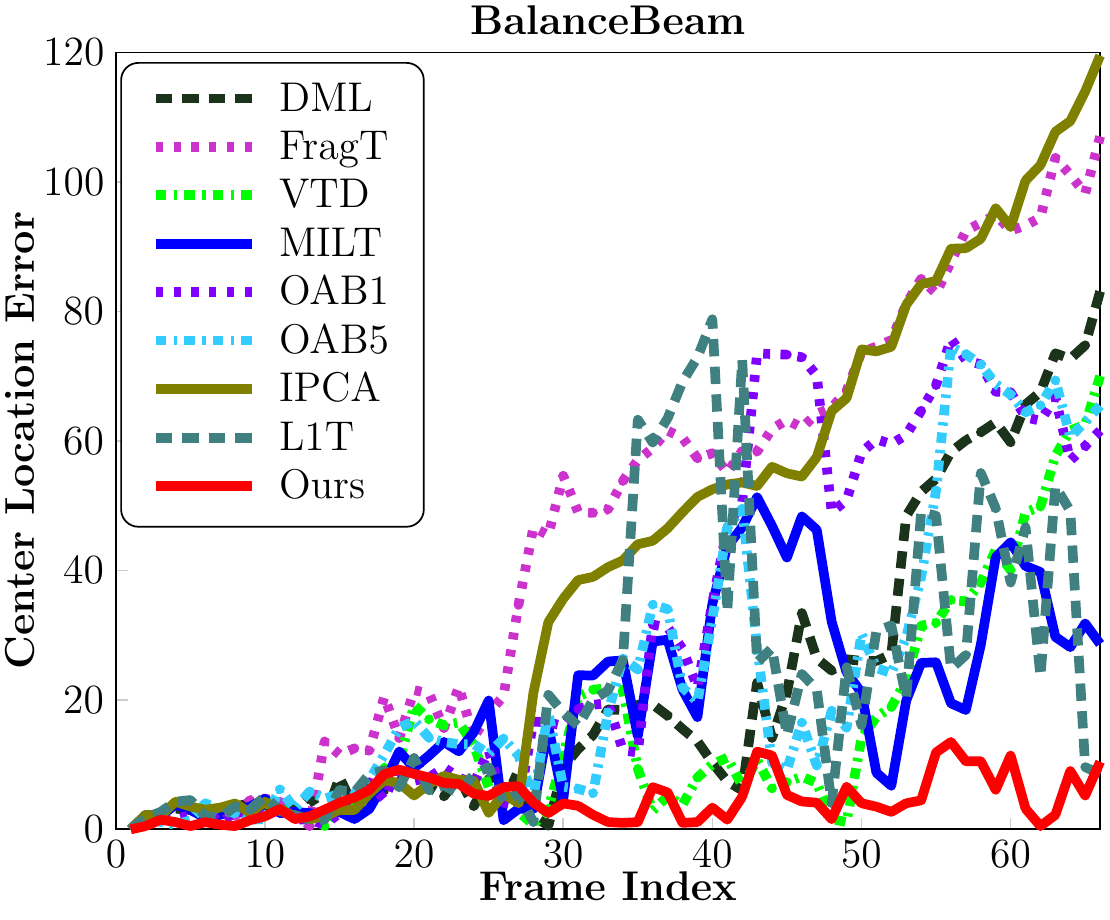}
\includegraphics[scale=0.3571]{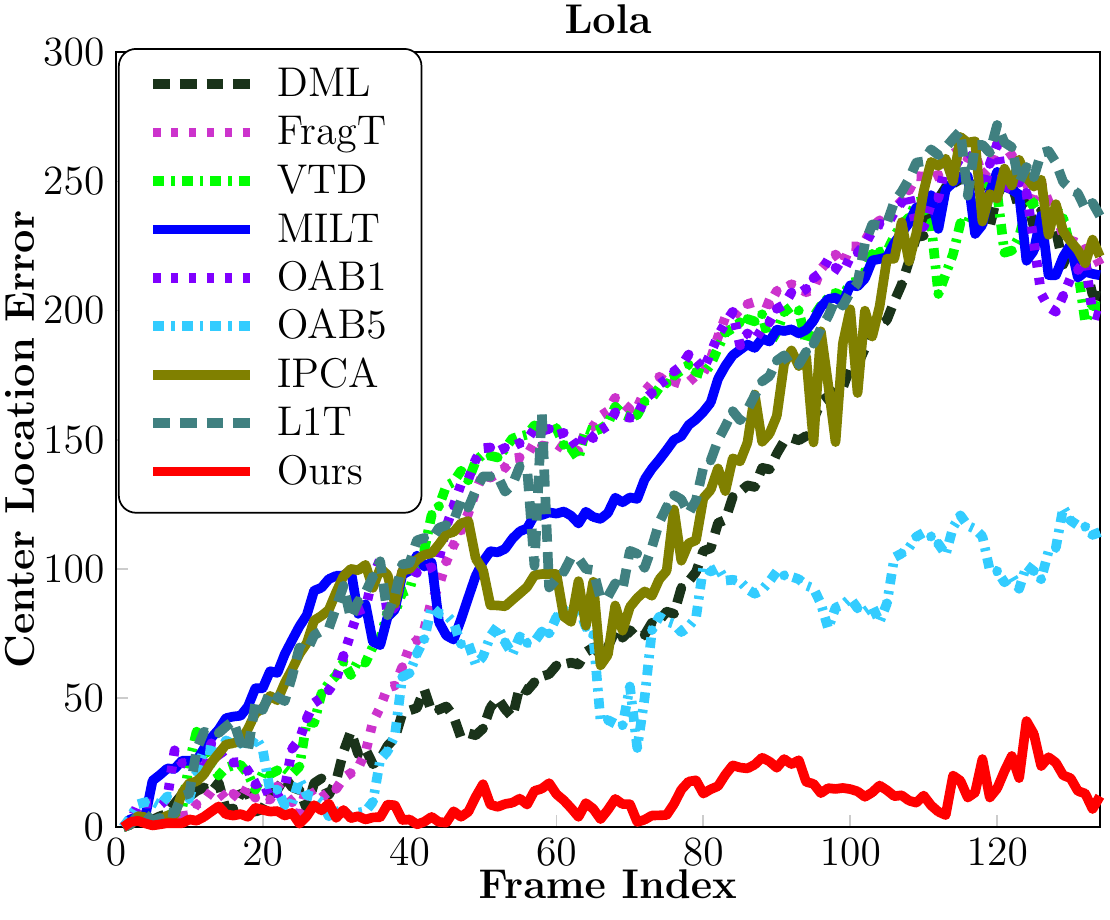}
\includegraphics[scale=0.3571]{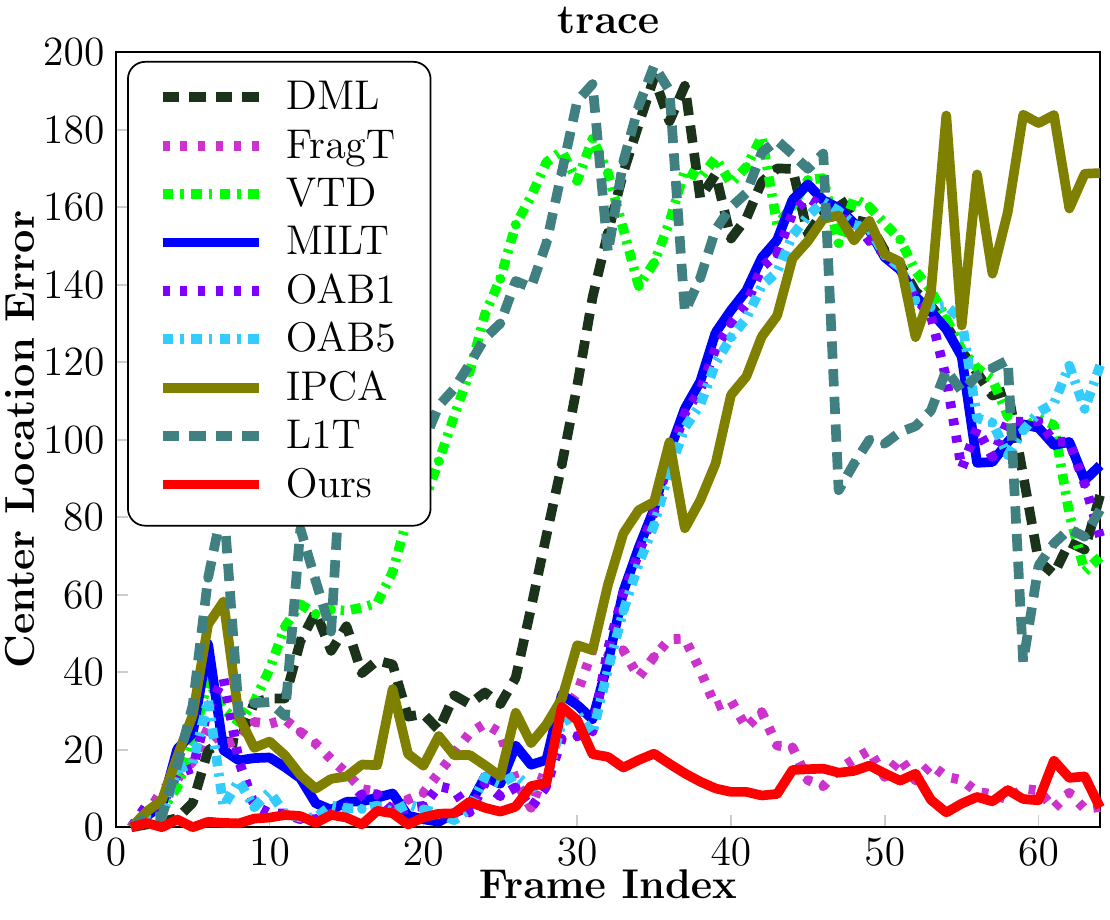}
\includegraphics[scale=0.3571]{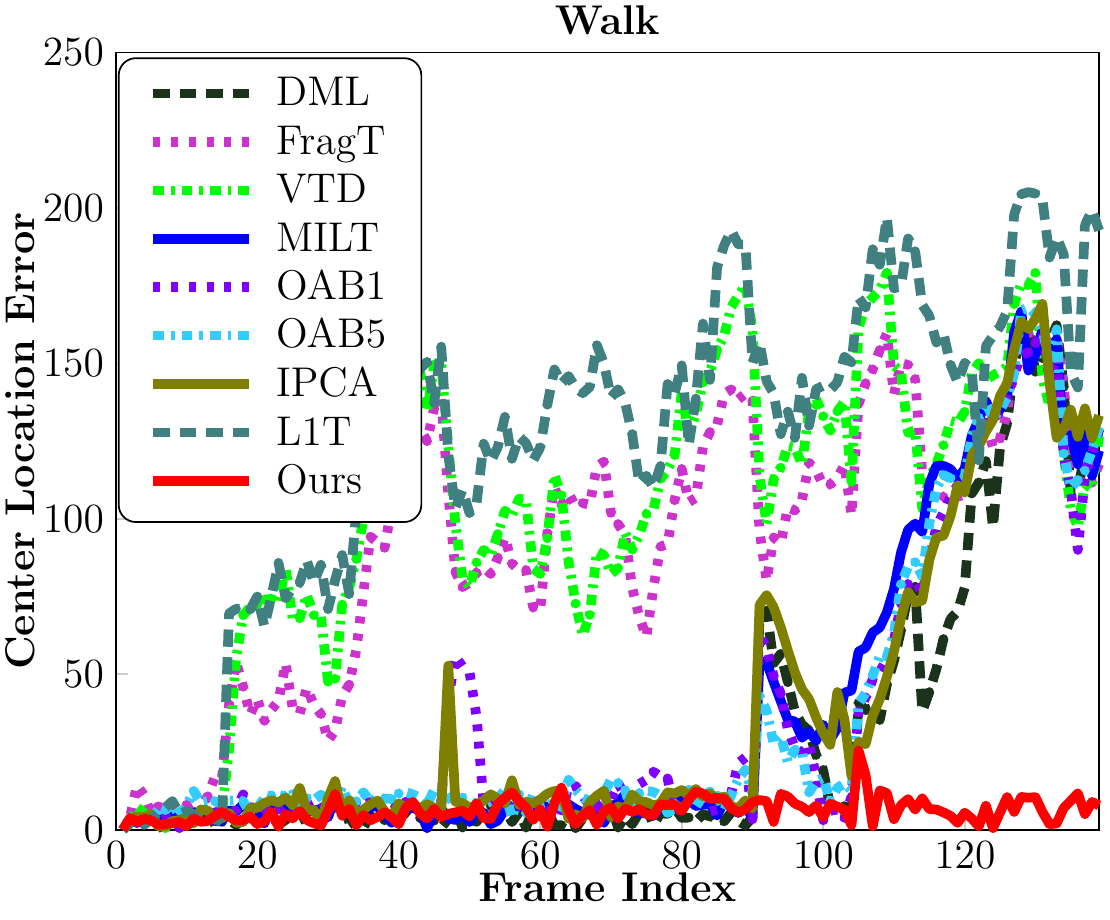}\\
\includegraphics[scale=0.3571]{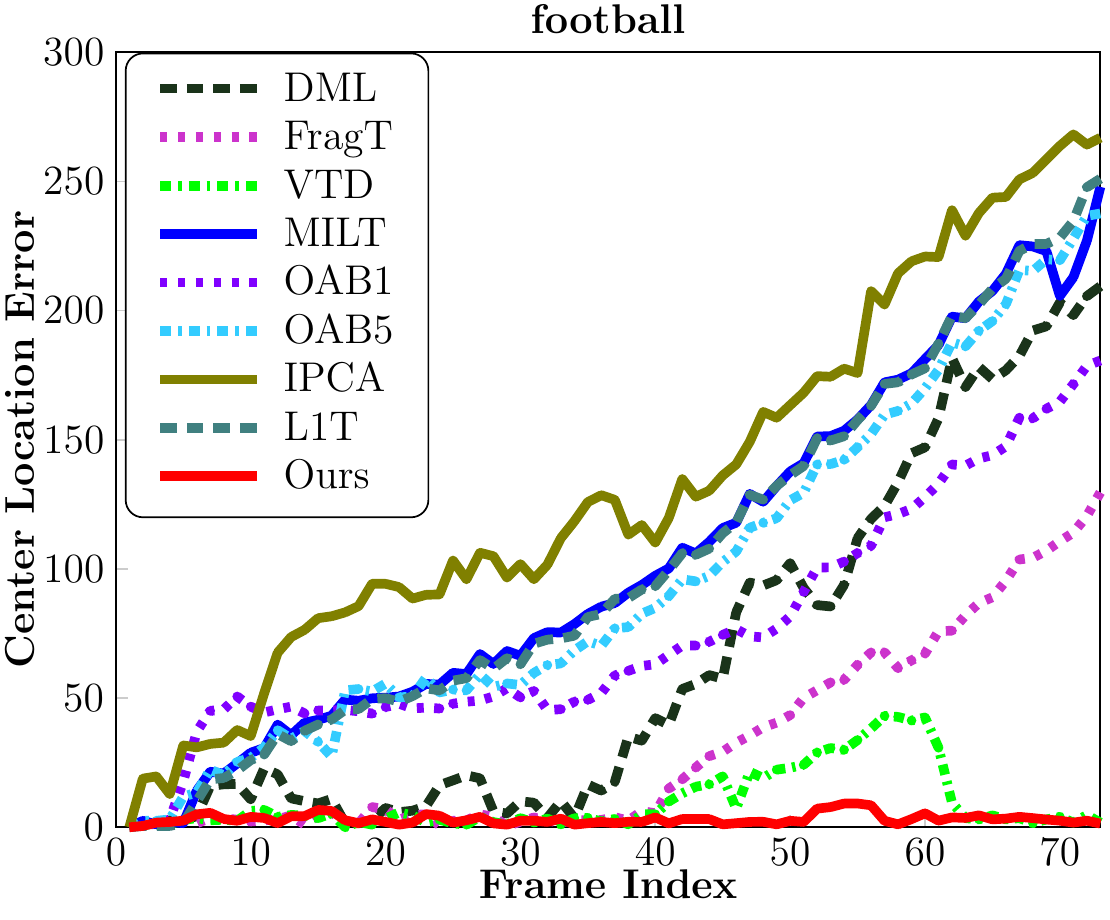}
\includegraphics[scale=0.3571]{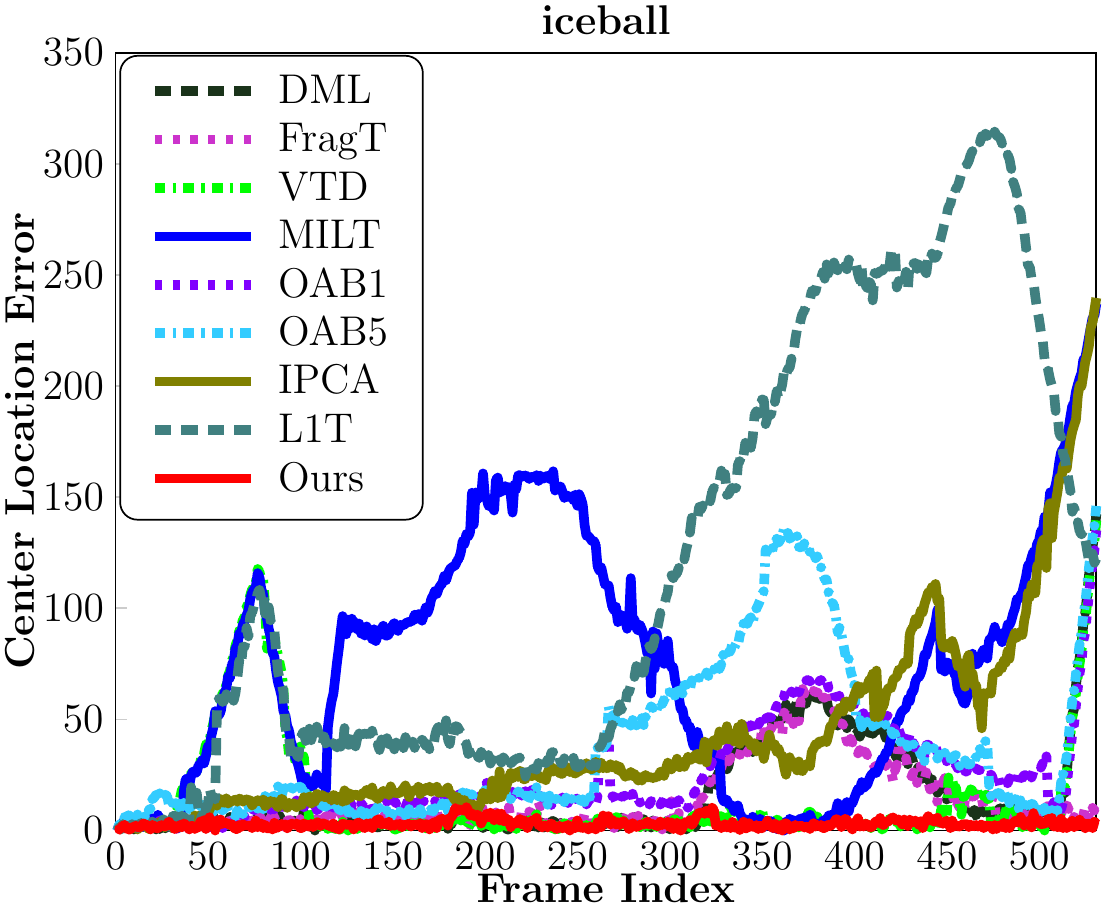}
\includegraphics[scale=0.3571]{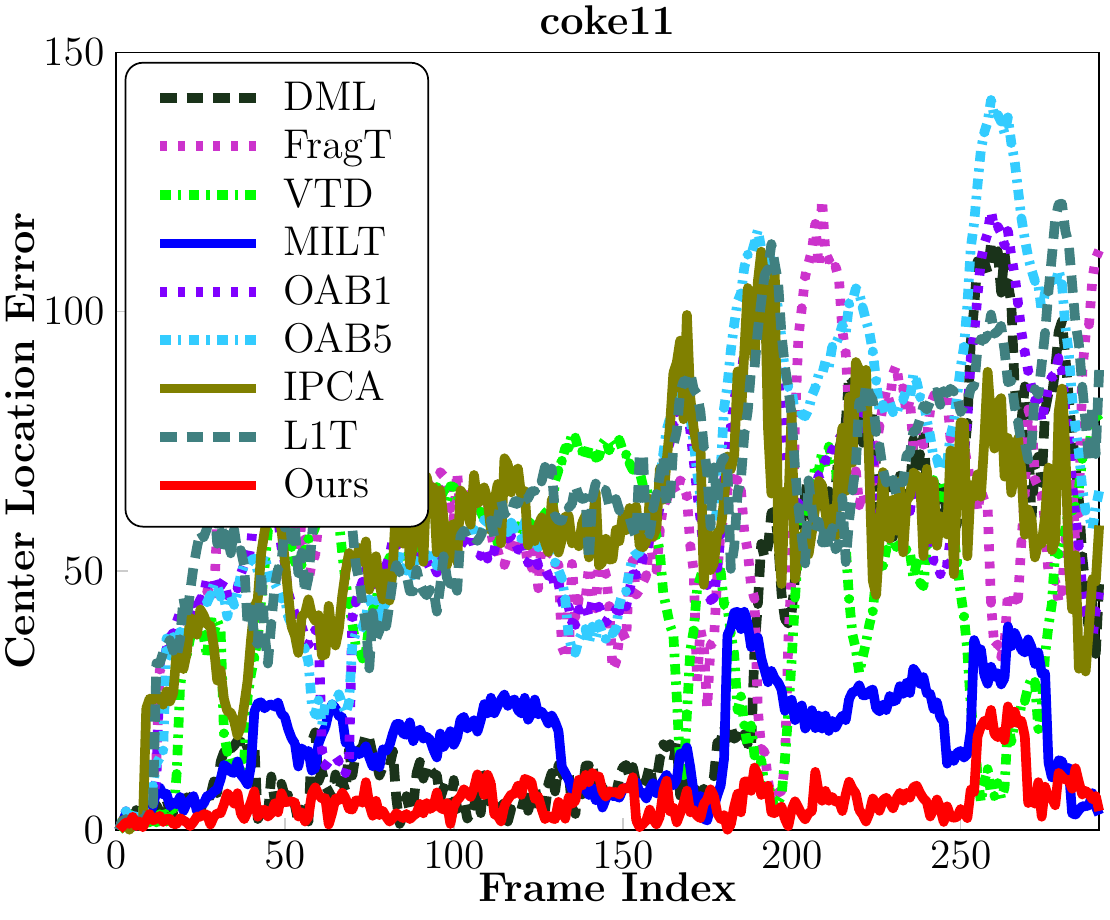}
\includegraphics[scale=0.3571]{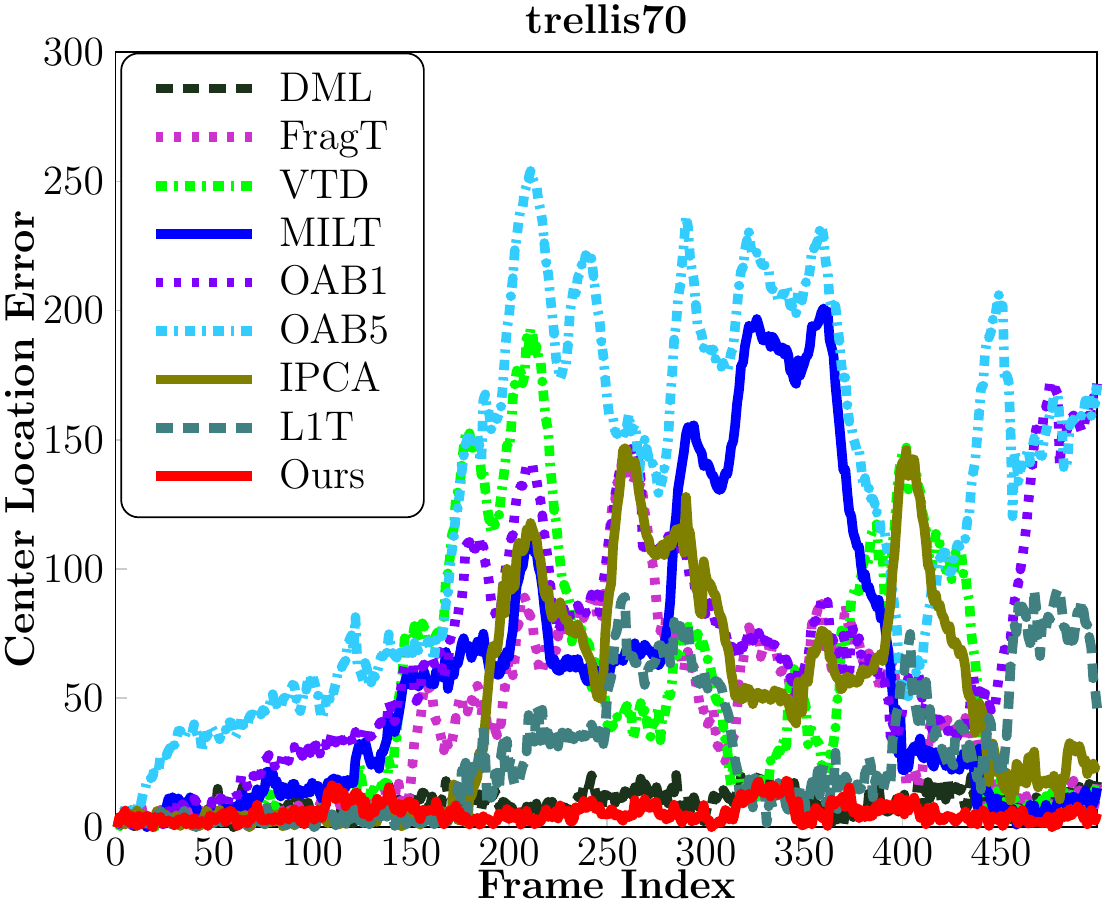}\\
\vspace{-0.33cm}
\caption{Quantitative comparison of different trackers in CLE on the first eight
video sequences.}
 \vspace{-0.46cm}
  \label{fig:exp_error_curve}
\end{figure*}

{\bf Comparison of competing trackers}
Fig.~\ref{fig:exp_error_curve} plots the frame-by-frame
center location errors (highlighted in different colors) obtained by
the nine trackers
for the first eight video sequences.
Tab.~\ref{Tab:quantitative} reports the success rates of the nine trackers
over the thirteen video sequences.
From Fig.~\ref{fig:exp_error_curve} and Tab.~\ref{Tab:quantitative},
we observe that the proposed tracking algorithm
achieves the best tracking performance on most video sequences.

{\bf Discussion}
Overall, the proposed tracking algorithm has the following properties. First,
after the buffer size exceeds a certain value (around $300$ in our
experiments), the tracking performance keeps stable with an increasing
buffer size, as shown in Fig.~\ref{fig:buffersize}. This is desirable since we do not need a large buffer
size to achieve promising performance.
Second, in contrast to many existing particle filtering-based trackers
whose running time is typically linear in the number of particles, our method's
running time is sublinear in the number of
particles, as shown in Fig.~\ref{fig:particle_number}.
Moreover, its tracking performance rapidly improves and
finally converge to a certain value, as shown in
Fig.~\ref{fig:particle_number}.
Third, as shown in Fig.~\ref{fig:metric_non_metric} and Tab.~\ref{Tab:metric_success_rate}, the performance of
our metric learning with
no eigendecomposition is close to that of computationally expensive metric learning with
step-by-step eigendecomposition.
Fourth, based on linear representation with metric learning, it performs
better in tracking accuracy, as shown
in
Fig.~\ref{fig:regression}. Fifth, it utilizes weighed reservoir
sampling to effectively maintain and update the foreground
and background sample buffers for metric learning, as shown in Fig.~\ref{fig:sampling}.
Last, compared with other state-of-the-art trackers, it is capable of
effectively
adapting to complicated appearance changes in the tracking process by
constructing
an effective metric-weighted linear representation with weighed
reservoir sampling, as shown in Fig.~\ref{fig:exp_error_curve} and Tab.~\ref{Tab:quantitative}.

\vspace{-0.1cm}
\section{Conclusion}
\vspace{-0.1cm}

    We have proposed a robust visual tracker based on
    non-sparse linear representations, which can be solved extremely
    efficiently in closed-form. Compared with recent sparse linear
    representation based trackers \cite{Meo-Ling-ICCV09,Li-Shen-Shi-cvpr2011},
    even with this simple implementation, our tracker is already much faster with
    comparable accuracy.
    To further improve the discriminative capacity of the linear
    representation, we have presented online Mahalanobis distance
    metric learning, which is able to capture the correlation
    information between feature dimensions. We empirically show that
    combining a metric into the linear representation considerably
    improve the robustness of the tracker. To make the online
    metric learning even more efficient, for the first time,
    we design a learning mechanism based on time-weighted
    reservoir sampling.
    With this mechanism, recently streamed samples in the video are
    assigned more importance weights.
    We have also theoretically proved that metric learning based on
    the proposed reservoir sampling with limited-sized sampling
    buffers can effectively approximate metric learning using all the
    received training samples.
    Compared with a few state-of-the-art trackers on thirteen challenging
    sequences, we empirically show that our method is  more robust to
    complicated appearance changes,  pose variations, and occlusions,
    etc.

\textbf{Acknowledgments}
This work is in part supported by ARC Discovery Project (DP1094764).

\begin{table}
\ninesevenhao
\begin{tabular}{l|c|c|c|c|c|c|c|c|c}
\hline
& \makebox[0.05cm]{\textbf{Ours}}   &  \makebox[0.2cm]{DML} &
\makebox[0.1cm]{FragT} & \makebox[0.1cm]{VTD} & \makebox[0.2cm]{MILT}
& \makebox[0.1cm]{OAB1} & \makebox[0.1cm]{OAB5}  &
\makebox[0.1cm]{IPCA} & \makebox[0.1cm]{L1T} \\
\hline
\hline

 \makebox[0.65cm]{B-Beam}  &  \bf \makebox[0.05cm]{0.94}  &
\makebox[0.25cm]{0.43}  &  \makebox[0.25cm]{0.25} &
\makebox[0.25cm]{0.56}  &  \makebox[0.25cm]{0.37}   &
\makebox[0.25cm]{0.37}  &  \makebox[0.25cm]{0.43}  &
\makebox[0.25cm]{0.34}  &\makebox[0.25cm]{0.43}\\
Lola &   \bf 0.80  &  0.18 &   0.11  &  0.07  &  0.03  &  0.01  &
0.08  &  0.06  &  0.07\\
trace   & \bf 0.89 &   0.12  &  0.63  &  0.11  &  0.42  &  0.40  &  0.42
 &  0.31  &  0.06\\
Walk   & \bf 0.98  &  0.67  &  0.09   & 0.11  &  0.64   & 0.62 &
0.67  &  0.62  &  0.11\\
football &  \bf 0.88  &  0.20  &  0.47   & 0.62  &  0.07   & 0.05  &
0.05 &   0.01   & 0.07\\
iceball  & \bf 0.93  &  0.59  &  0.52  &  0.70  &  0.16   & 0.14 &
0.12  &  0.09   & 0.08\\
coke11   & \bf 0.87  &  0.37  &  0.05   & 0.10  &  0.28  &  0.04 &
0.04  &  0.03   & 0.04\\

trellis70 &  \bf 0.98  &  0.90 &   0.40  &  0.37  &  0.34  &  0.13  &
0.03  &  0.38  &  0.34\\

dograce  & \bf 0.97  &  0.60  &  0.49  &  0.47   & 0.67   & 0.67  &
0.23   & 0.87   & 0.87\\
football3 &  \bf 0.97    &0.46  &  0.32  &  0.31  &  0.61 &   0.87   &
0.24   & 0.22   & 0.16\\
car11  &  \bf 0.99   & 0.92  &  0.08  &  0.43 &   0.08  &  0.39   &
0.33 &   \bf 0.99 &   0.59\\
cubicle &  \bf 0.98 &   0.82   & 0.37  &  0.20  &  0.22 &   0.22 &
0.21 &   0.25  &  0.49\\
seq-jd  & \bf 0.95   & 0.85  &  0.88  &  0.79  &  0.61 &   0.58 &
0.38  &  0.45 &   0.61\\
\hline
\end{tabular} \vspace{-0.27cm}
\caption{The quantitative comparison results of the nine trackers over
the thirteen video sequences.
The table reports their tracking success rates over each video sequence.
 }
\label{Tab:quantitative}
\vspace{-0.15cm}
\end{table}

{
\bibliography{ref}

\begin{thebibliography}{10}

\bibitem{Limy-Ross17}
D.~A. Ross, J.~Lim, R.~Lin, and M.~Yang,
\newblock ``Incremental learning for robust visual tracking,''
\newblock {\em Int. J. Comp. Vis.}, vol. 77, no. 1, pp. 125--141, 2008.

\bibitem{Meo-Ling-ICCV09}
X.~Mei and H.~Ling,
\newblock ``Robust visual tracking and vehicle classification via sparse
  representation,''
\newblock {\em {IEEE} Trans. Pattern Anal. Mach. Intell.}, 2011.

\bibitem{Kwon-Lee-CVPR2010}
J.~Kwon and K.~M. Lee,
\newblock ``Visual tracking decomposition,''
\newblock in {\em Proc. IEEE Conf. Comp. Vis. Patt. Recogn.}, 2010, pp.
  1269--1276.

\bibitem{Li-Shen-Shi-cvpr2011}
H.~Li, C.~Shen, and Q.~Shi,
\newblock ``Real-time visual tracking with compressive sensing,''
\newblock in {\em Proc. IEEE Conf. Comp. Vis. Patt. Recogn.}, 2011.

\bibitem{harestruck_iccv2011}
S.~Hare, A.~Saffari, and P.H.S. Torr,
\newblock ``Struck: Structured output tracking with kernels,''
\newblock in {\em Proc. IEEE Int. Conf. Comp. Vis.}, 2011.

\bibitem{li2007robust}
X.~Li, W.~Hu, Z.~Zhang, X.~Zhang, and G.~Luo,
\newblock ``Robust visual tracking based on incremental tensor subspace
  learning,''
\newblock in {\em Proc. IEEE Int. Conf. Comp. Vis.}, 2007, pp. 1--8.

\bibitem{li2011graph}
X.~Li, A.~Dick, H.~Wang, C.~Shen, and A.~{van den Hengel},
\newblock ``Graph mode-based contextual kernels for robust {SVM} tracking,''
\newblock in {\em Proc. IEEE Int. Conf. Comp. Vis.}, 2011, pp. 1156--1163.

\bibitem{li2010robust}
X.~Li, W.~Hu, H.~Wang, and Z.~Zhang,
\newblock ``Robust object tracking using a spatial pyramid heat kernel
  structural information representation,''
\newblock {\em Neurocomputing}, vol. 73, no. 16-18, pp. 3179--3190, 2010.

\bibitem{Generalized2010Shen}
C.~Shen, J.~Kim, and H.~Wang,
\newblock ``Generalized kernel-based visual tracking,''
\newblock {\em IEEE Trans. Circuits \& Systems for Video Tech.}, vol. 20, no.
  1, pp. 119--130, 2010.

\bibitem{Shi-Eriksson-Hengel-Shen-cvpr2011}
Q.~Shi, A.~Eriksson, A.~{van den Hengel}, and C.~Shen,
\newblock ``Is face recognition really a compressive sensing problem?,''
\newblock in {\em Proc. IEEE Conf. Comp. Vis. Patt. Recogn.}, 2011.

\bibitem{rigamonti2011sparse}
R.~Rigamonti, M.~A. Brown, and V.~Lepetit,
\newblock ``Are sparse representations really relevant for image
  classification?,''
\newblock in {\em Proc. IEEE Conf. Comp. Vis. Patt. Recogn.}, 2011, pp.
  1545--1552.

\bibitem{FaceICCV2011}
L.~Zhang, M.~Yang, and X.~Feng,
\newblock ``Sparse representation or collaborative representation: Which helps
  face recognition?,''
\newblock in {\em Proc. IEEE Int. Conf. Comp. Vis.}, 2011.

\bibitem{weinberger2006distance}
K.Q. Weinberger, J.~Blitzer, and L.K. Saul,
\newblock ``Distance metric learning for large margin nearest neighbor
  classification,''
\newblock in {\em Proc. Adv. Neural Inf. Process. Syst.}, 2006.

\bibitem{Shen2009PSD}
C.~Shen, J.~Kim, L.~Wang, and A.~{van den Hengel},
\newblock ``Positive semidefinite metric learning with boosting,''
\newblock in {\em Proc. Adv. Neural Inf. Process. Syst.}, 2009, pp. 1651--1659.

\bibitem{wang2010discriminative}
X.~Wang, G.~Hua, and T.~Han,
\newblock ``Discriminative tracking by metric learning,''
\newblock {\em Proc. Eur. Conf. Comp. Vis.}, pp. 200--214, 2010.

\bibitem{jiang2011adaptive}
N.~Jiang, W.~Liu, and Y.~Wu,
\newblock ``Adaptive and discriminative metric differential tracking,''
\newblock in {\em Proc. IEEE Conf. Comp. Vis. Patt. Recogn.}, 2011, pp.
  1161--1168.

\bibitem{chechik2010large}
G.~Chechik, V.~Sharma, U.~Shalit, and S.~Bengio,
\newblock ``Large scale online learning of image similarity through ranking,''
\newblock {\em J. Mach. Learn. Research}, vol. 11, pp. 1109--1135, 2010.

\bibitem{vitter1985random}
J.~S. Vitter,
\newblock ``Random sampling with a reservoir,''
\newblock {\em ACM Trans. Math. Software}, vol. 11, no. 1, pp. 37--57, 1985.

\bibitem{zhaoICML2011}
P.~Zhao, S.C.H. Hoi, R.~Jin, and T.~Yang,
\newblock ``Online {AUC} maximization,''
\newblock in {\em Proc. Int. Conf. Mach. Learn.}, 2011.

\bibitem{Kolonko04sequentialreservoir}
M.~Kolonko and D.~W\"asch,
\newblock ``Sequential reservoir sampling with a non-uniform distribution,''
\newblock {\em {ACM} Trans. Math. Software}, vol. 32, pp. 257--273, 2004.

\bibitem{efraimidis2006weighted}
P.~S. Efraimidis and P.~G. Spirakis,
\newblock ``Weighted random sampling with a reservoir,''
\newblock {\em Information process. letters}, vol. 97, no. 5, pp. 181--185,
  2006.

\bibitem{jennings1992matrix}
A.~Jennings and J.~McKeown,
\newblock {\em Matrix computation},
\newblock John Wiley \& Sons Inc., 1992.

\bibitem{householder1964theory}
A.~S. Householder,
\newblock {\em The theory of matrices in numerical analysis},
\newblock Blaisdell Publishing Co.: New York, 1964.

\bibitem{powell1969theorem}
M.~J.~D. Powell,
\newblock ``A theorem on rank one modifications to a matrix and its inverse,''
\newblock {\em The Computer Journal}, vol. 12, no. 3, pp. 288--290, 1969.

\bibitem{crammer2006online}
K.~Crammer, O.~Dekel, J.~Keshet, S.~Shalev-Shwartz, and Y.~Singer,
\newblock ``Online passive-aggressive algorithms,''
\newblock {\em J. Mach. Learn. Research}, vol. 7, pp. 551--585, 2006.

\bibitem{Dalal-Triggs-CVPR2005}
N.~Dalal and B.~Triggs,
\newblock ``Histograms of oriented gradients for human detection,''
\newblock in {\em Proc. IEEE Conf. Comp. Vis. Patt. Recogn.}, 2005.

\bibitem{lixi-cvpr2008}
X.~Li, W.~Hu, Z.~Zhang, X.~Zhang, M.~Zhu, and J.~Cheng,
\newblock ``Visual tracking via incremental log-euclidean riemannian subspace
  learning,''
\newblock in {\em Proc. IEEE Conf. Comp. Vis. Patt. Recogn.}, 2008, pp. 1--8.

\bibitem{Adam-Fragment-2006}
A.~Adam, E.~Rivlin, and I.~Shimshoni,
\newblock ``Robust fragments-based tracking using the integral histogram,''
\newblock in {\em Proc. IEEE Conf. Comp. Vis. Patt. Recogn.}, 2006, pp.
  798--805.

\bibitem{Babenko-Yang-Belongie-cvpr2009}
B.~Babenko, M.~Yang, and S.~Belongie,
\newblock ``Visual tracking with online multiple instance learning,''
\newblock in {\em Proc. IEEE Conf. Comp. Vis. Patt. Recogn.}, 2009, pp.
  983--990.

\bibitem{Grabner-Grabner-Bischof-BMVC2006}
H.~Grabner, M.~Grabner, and H.~Bischof,
\newblock ``Real-time tracking via on-line boosting,''
\newblock in {\em Proc. British Machine Vis. Conf.}, 2006, pp. 47--56.

\end{thebibliography}
\bibliographystyle{ieee}
}
\end{document}